\documentclass[10pt,journal,cspaper,compsoc]{IEEEtran}
\usepackage{epsfig}
\usepackage{graphicx}
\usepackage{algorithm}
\usepackage{algorithmic}
\usepackage{amsmath}
\usepackage{amssymb}
\usepackage{mathrsfs} 
\usepackage{amsmath,bm}
\usepackage{array,multirow}
\usepackage{mdwlist}
\usepackage{paralist}
\usepackage{url}
\usepackage{color}
\usepackage{caption}
\usepackage{setspace}
\usepackage{textcomp}
\usepackage{bbding}
\usepackage{booktabs}
\usepackage{threeparttable}
\usepackage{subfigure}
\usepackage[colorlinks=true,linkcolor=blue,anchorcolor=blue,citecolor=blue]{hyperref}

\def\ie{{\em i.e.}}
\def\eg{{\em e.g.}}
\def\etal{{\em et al.}}

\ifCLASSOPTIONcompsoc
  \usepackage[nocompress]{cite}
\else
  \usepackage{cite}
\fi

\ifCLASSINFOpdf
\else
\fi

\begin{document}
%
\title{Detection and Tracking Meet Drones Challenge}

\author{Pengfei Zhu$^\ast$, Longyin Wen$^\ast$, Dawei Du$^\ast$, Xiao Bian, Heng Fan, Qinghua Hu, Haibin Ling
\IEEEcompsocitemizethanks{\IEEEcompsocthanksitem Pengfei Zhu, Longyin Wen, and Dawei Du contributed equally to this work. The order of names is determined by coin flipping.
\IEEEcompsocthanksitem Pengfei Zhu and Qinghua Hu are with the College of Intelligence and Computing, Tianjin University, Tianjin, China (e-mail: \{zhupengfei, huqinghua\}@tju.edu.cn).
\IEEEcompsocthanksitem Longyin Wen is with the JD Finance America Corporation, Mountain View, CA, USA (e-mail: longyin.wen@jd.com).
\IEEEcompsocthanksitem Dawei Du is with the Computer Science Department, University at Albany, State University of New York, Albany, NY, USA (e-mail: ddu@albany.edu).
\IEEEcompsocthanksitem Xiao Bian is with GE Global Research, Niskayuna, NY, USA (e-mail: xiao.bian@ge.com).
\IEEEcompsocthanksitem Heng Fan is with the Department of Computer Science and Engineering, University of North Texas, Denton, TX, USA (e-mail: heng.fan@unt.edu).
\IEEEcompsocthanksitem Haibin Ling is with the Department of Computer Science, Stony Brook University, New York, NY, USA (e-mail: hling@cs.stonybrook.edu).}
}

\markboth{}%
{P. Zhu, L. Wen and D. Du {\textit{et al.}}: Detection and Tracking Meet Drones Challenge}

\def \VIS {VisDrone}

\IEEEtitleabstractindextext{%
\begin{abstract}
Drones, or general UAVs, equipped with cameras have been fast deployed with a wide range of applications, including agriculture, aerial photography, and surveillance. Consequently, automatic understanding of visual data collected from drones becomes highly demanding, bringing computer vision and drones more and more closely. To promote and track the developments of object detection and tracking algorithms, we have organized three challenge workshops in conjunction with ECCV 2018, ICCV 2019 and ECCV 2020, attracting more than $100$ teams around the world. We provide a large-scale drone captured dataset, \VIS, which includes four tracks, \ie, (1) image object detection, (2) video object detection, (3) single object tracking, and (4) multi-object tracking. In this paper, we first present a thorough review of object detection and tracking datasets and benchmarks, and discuss the challenges of collecting large-scale drone-based object detection and tracking datasets with fully manual annotations. After that, we describe our \VIS~dataset, which is captured over various urban/suburban areas of $14$ different cities across China from North to South. Being the largest such dataset ever published, \VIS~enables extensive evaluation and investigation of visual analysis algorithms for the drone platform. We provide a detailed analysis of the current state of the field of large-scale object detection and tracking on drones, and conclude the challenge as well as propose future directions. We expect the benchmark largely boost the research and development in video analysis on drone platforms. All the datasets and experimental results can be downloaded from \url{https://github.com/VisDrone/VisDrone-Dataset}.
\end{abstract}

}

\maketitle

\IEEEdisplaynontitleabstractindextext

\IEEEpeerreviewmaketitle

\IEEEraisesectionheading{\section{Introduction}\label{sec:introduction}}
\IEEEPARstart{D}{rones} (or UAVs) equipped with cameras are receiving a lot of attention in recent years. 
The commercial drone market report \cite{drone-report} describes that the global commercial drone market size will reach $501.4$ billion by 2028, with a compound annual growth rate of $57.5\%$ from 2021 to 2028. Equipped with embedded devices, drones are able to analyze the captured data and spawn a variety of new application scenarios, \eg,
\begin{itemize}
\item \textbf{Agriculture}. Drones can provide valuable insights to help farmers or ranchers optimize agriculture operations, monitor crop growth and keep herds safe, etc.
\item \textbf{Aerial photography}. Drones are used to extract aerial photography images instead of expensive cranes and helicopters.
\item \textbf{Shipping and delivery}. Drones can efficiently send packages such as medical supplies, food, or other goods to the designated places.
\item \textbf{Security and surveillance}. Drones can provide real-time visibility into security threats and emergency situations by monitoring large regions.
\item \textbf{Search and rescue}. Drones are useful to help search missing persons, fugitives, or rescue survivors and drop supplies in difficult terrains and harsh conditions.
\end{itemize}

Consequently, automatic understanding of visual data collected from drones become highly demanding, which brings computer vision to drones more and more closely. As two fundamental problems in computer vision, object detection and object tracking are under extensive investigation. However, although the great progress has been made in various application scenarios such as Internet and security surveillance, they are not usually optimal for dealing with drone captured sequences or images. Notably, comparing to existing datasets in computer vision field, the drone-captured video sequences bring several new challenges, described as follows. 
\begin{itemize}
\item {\em Viewpoint variations}: comparing to surveillance cameras with fixed viewpoints, drone-equipped cameras monitor the objects in arbitrary viewpoints \cite{DBLP:conf/eccv/MuellerSG16}. 
\item {\em Scale variations}: drone-equipped cameras monitor the objects at different altitudes, resulting in large variations of scales of objects \cite{DBLP:conf/fgr/KalraSN0VS19}. 
\item {\em Motion blur}: videos are generally recorded by the drone-equipped cameras in the moving process, bringing in considerable motion blurs of the recorded videos \cite{DBLP:conf/eccv/DuQYYDLZHT18}. 
\end{itemize}

Thus, it is necessary to develop and evaluate new vision algorithms for drone captured visual data. However, as pointed out in \cite{DBLP:conf/eccv/MuellerSG16,DBLP:conf/iccv/HsiehLH17}, studies toward this goal are seriously limited by the lack of publicly available large-scale benchmarks or datasets. Some recent efforts \cite{DBLP:conf/eccv/MuellerSG16,DBLP:conf/eccv/RobicquetSAS16,DBLP:conf/iccv/HsiehLH17} have been devoted to construct datasets captured by drones focusing on object detection or tracking. These datasets are still limited in size and scenarios covered, due to the difficulties in data collection and annotation. Thorough evaluations of existing or newly developed algorithms remain an open problem. A more general and comprehensive benchmark is desired for further boosting video analysis research on drone platforms.

Thus motivated, we have organized three challenge workshops in conjunction with European Conference on Computer Vision (ECCV) 2018 \cite{DBLP:conf/eccv/ZhuWDBLHNCLLMWW18,DBLP:conf/eccv/WenZDBLHLCLMNWW18,DBLP:conf/eccv/ZhuWDBLHWNCLLMW18}, IEEE International Conference on Computer Vision (ICCV) 2019 \cite{DBLP:conf/iccv/DuZWBLHPDET19,DBLP:conf/iccv/DuZWBLHZPWZBS19,DBLP:conf/iccv/ZhuDWBLHPZWZBS19,DBLP:conf/iccv/WenZDBLHZPWZBS19}, and European Conference on Computer Vision (ECCV) 2020 \cite{DBLP:conf/eccv/DuWZFHLSPASPJDL20,DBLP:conf/eccv/FanDWZHLSPSDSXB20,DBLP:conf/eccv/FanWDZHLSWDYWZS20}, attracting more than $100$ research teams around the world. The challenge focuses on object detection and tracking with four tracks.
\begin{itemize}
\item {\em Image object detection track (DET).} Given a pre-defined set of object classes, \eg, cars and pedestrians, the algorithm is required to detect objects of these classes from individual images taken by drones.
\item {\em Video object detection track (VID).} Similar to DET, the algorithm is required to detect objects of predefined object classes from videos taken by drones.
\item {\em Single object tracking track (SOT).} The goal of the track is to estimate the state of a target, indicated in the first frame, across frames in an online manner.
\item {\em Multi-object tracking track (MOT).} This track aims to localize object instances in each video frame and recover their trajectories in video sequences.
\end{itemize}

Notably, in the workshop challenges, we provide a large-scale dataset, which consists of $263$ video clips with $179,264$ frames and $10,209$ static images. The data is recorded by various drone-mounted cameras, diverse in many aspects including location (taken from $14$ different cities in China), environment (urban and rural regions), objects (\eg, pedestrian, vehicles, and bicycles), and density (sparse and crowded scenes). We select $10$ categories of objects of frequent interests in drone applications, such as {\em pedestrians} and {\em cars}. Altogether we carefully annotate more than $2.5$ million bounding boxes of object instances from these categories. Moreover, some important attributes including visibility of the scenes, object category and occlusion, are provided for better data usage. The detailed comparison of the provided drone datasets with other related datasets in object detection and tracking are presented in Table \ref{tab:comparison-dataset}.

{\noindent {\bf Scope of this paper}.}
This paper summarizes the \VIS-Challenges organized from 2018 to 2020, which is extended from the challenge summary papers \cite{DBLP:conf/eccv/ZhuWDBLHNCLLMWW18,DBLP:conf/iccv/DuZWBLHPDET19,DBLP:conf/eccv/DuWZFHLSPASPJDL20,DBLP:conf/eccv/WenZDBLHLCLMNWW18,DBLP:conf/iccv/DuZWBLHZPWZBS19,DBLP:conf/eccv/FanWDZHLSWDYWZS20,DBLP:conf/eccv/ZhuWDBLHWNCLLMW18,DBLP:conf/iccv/ZhuDWBLHPZWZBS19,DBLP:conf/iccv/WenZDBLHZPWZBS19,DBLP:conf/eccv/FanDWZHLSPSDSXB20}. Note that the previous summary papers just briefly introduce the \VIS~dataset, enumerate the submitted methods, and report the evaluation results. In contrast, this paper focuses on analyzing various representative object detection and tracking algorithms thoroughly. Specifically, we discuss and analyze the advantages and disadvantages of the submitted methods in terms of model design and training strategies. After that, we advocate several future research directions of object detection and tracking on drone-captured videos. We expect such comprehensive review and analysis to largely boost the research and development in video analysis on drones. In summary, the focus of this paper is twofold, \ie, 
\begin{itemize}
\item A new large-scale benchmark captured by drone-equipped cameras is proposed for DET, VID, SOT and MOT, which provides a comprehensive evaluation platform for object detection and tracking.
\item We discuss the technical trend based on the state-of-the-art methods and forecast several potential research directions in this field.
\end{itemize}

\begin{table*}[t]
\centering
\caption{Comparison of the existing benchmarks and datasets. Note that, the resolution indicates the maximum resolution of videos/images included in the benchmarks and datasets. ($1k=1,000$)}
\label{tab:comparison-dataset}
  \setlength{\tabcolsep}{8.0pt}
\scriptsize{
\begin{tabular}{|c|c|c|c|c|c|c|c|}
\hline
{\bf Image object detection}                                                                          &scenario  &\#images  &categories &avg. \#labels/categories &resolution &occlusion labels &year \\
\hline
PASCAL VOC2012 \cite{pascal-voc-2012}                                                         &life &$22.5k$ &$20$ &$1,373$ &$469\times387$ &$\surd$ &2012 \\
ImageNet Object Detection \cite{DBLP:journals/ijcv/RussakovskyDSKS15}      &life &$456.2k$ &$200$ &$2,007$ &$482\times415$ &$\surd$ &2013 \\
MS COCO \cite{DBLP:conf/eccv/LinMBHPRDZ14}                                            &life &$328.0k$ &$91$ &$27.5k$ &$640\times640$ & &2014 \\
COWC \cite{DBLP:conf/eccv/MundhenkKSB16}                                                &aerial   &$32.7k$  &$1$ &$32.7k$ &$2048\times2048$ & &2016\\
CARPK \cite{DBLP:conf/iccv/HsiehLH17}                                                          &drone &$1,448$ &$1$ &$89.8k$ &$1280\times720$ & &2017 \\
DOTA \cite{DBLP:conf/cvpr/XiaBDZBLDPZ18} &aerial &$2,806$ &$15$ &$12.6k$ &$12029\times5014$ & &2018 \\
MOR-UAV \cite{DBLP:conf/mm/MandalKV20} &drone &$10,948$ &$2$ &$44.9k$ &$1920\times1080$ & &2020 \\
{\bf \VIS}                                                                                              &drone &$10,209$ &$10$ &$54.2k$ &$2000\times1500$
 &$\surd$ &2018 \\
\hline
\end{tabular}}

\begin{tabular}{c}
\\
\end{tabular}

  \setlength{\tabcolsep}{7.5pt}
\scriptsize{
\begin{tabular}{|c|c|c|c|c|c|c|c|}
\hline
{\bf Video object detection} &scenario &\#frames &categories &avg. \#labels/categories &resolution &occlusion labels &year \\
\hline
ImageNet Video Detection \cite{DBLP:journals/ijcv/RussakovskyDSKS15}      &life &$2017.6k$ &$30$ &$66.8k$ &$1280\times1080$ &$\surd$ &2015 \\
UA-DETRAC Detection \cite{DBLP:journals/corr/WenDCLCQLYL15}                &surveillance &$140.1k$ &$4$ &$302.5k$ &$960\times540$ &$\surd$ &2015 \\
MOT17Det \cite{DBLP:journals/ijcv/DendorferOMSCRR21}      &life &$11.2k$ &$1$ &$392.8k$ &$1920\times1080$ &$\surd$ &2017 \\
Okutama-Action \cite{DBLP:conf/cvpr/BarekatainMSMNM17}      &drone &$77.4k$ &$1$ &$422.1k$ &$3840\times2160$ & &2017 \\
UAVDT-DET \cite{DBLP:conf/eccv/DuQYYDLZHT18}      &drone &$40.7k$ &$3$ &$267.6k$ &$1080\times540$ &$\surd$ &2018 \\
DroneSURF \cite{DBLP:conf/fgr/KalraSN0VS19} &drone &$411.5k$ &$1$ &$786.8k$ &$1280\times720$ & &2019 \\
{\bf \VIS}                                                                                             &drone &$40.0k$ &$10$ &$183.3k$ &$3840\times2160$ &$\surd$ &2018 \\
\hline
\end{tabular}}

\begin{tabular}{c}
\\
\end{tabular}

  \setlength{\tabcolsep}{31.5pt}
\scriptsize{
\begin{tabular}{|c|c|c|c|c|}
\hline
{\bf Single object tracking}                                               &scenarios &\#sequences &\#frames &year \\
\hline

OTB100 \cite{DBLP:journals/pami/WuLY15}                      &life &$100$ &$59.0k$ &2015 \\
VOT2016 \cite{DBLP:conf/eccv/KristanLMFPCVHL16}     &life &$60$ &$21.5k$ &2016 \\
UAV123 \cite{DBLP:conf/eccv/MuellerSG16}                    &drone &$123$ &$110k$ &2016 \\
DTB70 \cite{DBLP:conf/aaai/LiY17}                    &drone &$70$ &$15.8k$ &2017 \\
UAVDT-SOT \cite{DBLP:conf/eccv/DuQYYDLZHT18}  &drone &$50$ &$37.2k$ &2018 \\
GOT-10k~\cite{DBLP:journals/corr/abs-1810-11981} &life &$10.0k$ &$1500.0k$ &2018\\
TrackingNet~\cite{DBLP:conf/eccv/MullerBGAG18} &life &$30.6k$ &$1443.1k$ &2018\\
MDOT~\cite{DBLP:journals/corr/abs-2003-06994} &drone &$373$ &$259.8k$ &2020 \\
LaSOT~\cite{DBLP:journals/corr/abs-1809-07845} &life &$1.55k$ &$3870.0k$ &2021 \\
Anti-UAV~\cite{DBLP:journals/corr/abs-2101-08466} &drone &$318$ &$585.9k$ &2021 \\
{\bf \VIS}                                        &drone &$167$ &$139.3k$ &2018 \\
\hline
\end{tabular}}

\begin{tabular}{c}
\\
\end{tabular}

  \setlength{\tabcolsep}{8.3pt}
\scriptsize{
\begin{tabular}{|c|c|c|c|c|c|c|c|}
\hline
{\bf Multi-object tracking} &scenario &\#frames &categories &avg. \#labels/categories &resolution &occlusion labels &year \\
\hline
KITTI Tracking \cite{DBLP:conf/cvpr/GeigerLU12}   &driving &$19.1k$ &$5$ &$19.0k$ &$1392\times512$  &$\surd$ &2013 \\
MOT15 \cite{DBLP:journals/ijcv/DendorferOMSCRR21} &surveillance &$11.3k$ &$1$ &$101.3k$ &$1920\times1080$ & &2015 \\
UA-DETRAC Tracking \cite{DBLP:journals/corr/WenDCLCQLYL15}   &surveillance &$140.1k$ &$4$ &$302.5k$ &$960\times540$ &$\surd$ &2015 \\
DukeMTMC \cite{DBLP:conf/eccv/RistaniSZCT16}      &surveillance &$2852.2k$ &$1$ &$4077.1k$ &$1920\times1080$ & &2016 \\
Campus \cite{DBLP:conf/eccv/RobicquetSAS16}      &drone &$929.5k$ &$6$ &$1769.4k$ &$1417\times2019$ & &2016\\
MOT17 \cite{DBLP:journals/ijcv/DendorferOMSCRR21}  &surveillance &$11.2k$ &$1$ &$392.8k$ &$1920\times1080$ & &2017 \\
UAVDT-MOT \cite{DBLP:conf/eccv/DuQYYDLZHT18} &drone &$40.7k$ &$3$ &$267.6k$ &$1080\times540$ &$\surd$ &2018 \\
TAO \cite{DBLP:conf/eccv/DaveKTSR20} &life &$2674.4k$ &$833$ &$0.4k$ &$1280\times720$ & &2020 \\
{\bf \VIS}      &drone &$40.0k$ &$10$ &$183.3k$ &$3840\times2160$ &$\surd$ &2018 \\
\hline
\end{tabular}}
\label{tab:comparison-datasets}
\end{table*}

\section{Related Work}
\begin{figure*}[t]
\centering
\includegraphics[width=.95\linewidth]{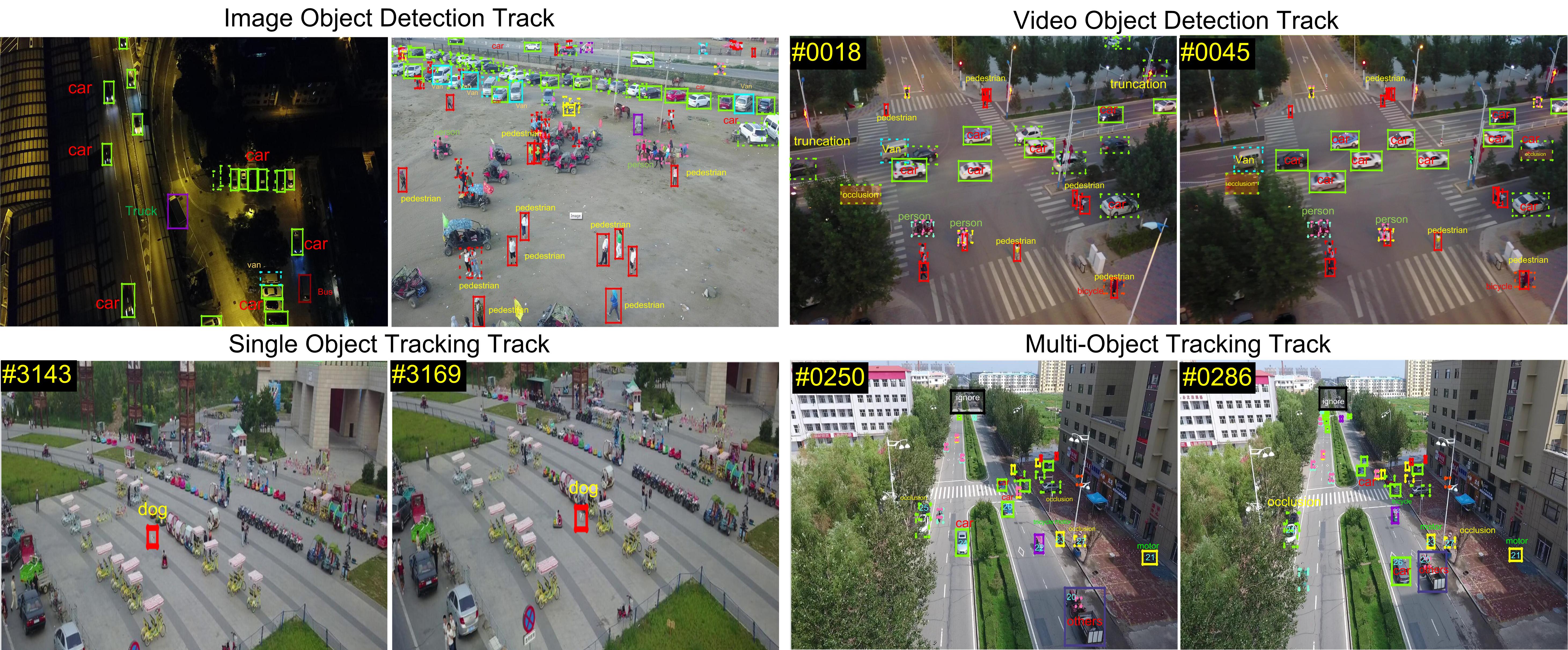}
\caption{Annotated example images in the proposed datasets. The dashed bounding box indicates the object is occluded. Different colors indicate different classes of objects. For better visualization, only a few attributes are displayed. }
\label{fig:visdrone_example}
\end{figure*}

\subsection{Surveys and Related Challenges}
Several survey papers are presented to discuss different topics in computer vision, such as generic object detection \cite{DBLP:journals/ijcv/LiuOWFCLP20}, single object tracking \cite{DBLP:journals/corr/abs-1912-00535}, and multi-object tracking \cite{DBLP:journals/ijon/CiaparroneSTTTH20}. Liu \etal \cite{DBLP:journals/ijcv/LiuOWFCLP20} summarize more than $300$ papers on generic object detection in terms of the detection framework, object feature extraction, object proposal generation, context modeling, training strategies, etc. In \cite{DBLP:journals/corr/abs-1912-00535}, the advantages and disadvantages of recent deep learning based trackers are investigated. Ciaparrone \etal \cite{DBLP:journals/ijon/CiaparroneSTTTH20} identify four main steps in multi-object tracking algorithms and present an in-depth review of the deep learning algorithms employed in each stage. In contrast to previous surveys solely focusing on generic object detection or tracking, we review and analyze the algorithms for object detection and tracking on drone-captured videos, which is useful to provide researchers a reference and guidance for algorithm design. 

In recent years, several challenges are organized to promote the developments of algorithms in object detection and tracking. The international workshop on computer vision for UAVs\footnote{\url{https://sites.google.com/site/uavision2018/}.} focuses on hardware, software and algorithmic (co-)optimizations towards the state-of-the-art image processing strategies on UAVs. Similarly, the Lower Power Object Detection Challenge \cite{DBLP:journals/pami/XuZYHRHS21} focuses on designing and implementing efficient object detection methods for UAVs. The VOT workshop\footnote{\url{http://www.votchallenge.net/}.} provides the tracking community with a precisely defined and repeatable way to compare short-term trackers as well as provides a common platform for discussing the evaluation and advancements made in the field of single-object tracking. The BMTT and BMTT-PETS workshops\footnote{\url{https://motchallenge.net/}.} aims to pave the way for a unified framework towards more meaningful evaluation of multi-object tracking. The Tiny Object Detection challenge\footnote{\url{https://rlq-tod.github.io/index.html}.} opens an interesting but challenging task, \ie, tiny person detection in unconstrained environments. Different from the aforementioned challenges and workshops, our workshop challenge aims to promote the developments of object detection and tracking on drone-captured videos, formed by four tracks, \ie, DET, VID, SOT, and MOT. 

\subsection{Object Detection and Tracking Datasets}
{\noindent {\bf Image object detection datasets.}} 
There are several image object detection datasets \cite{DBLP:conf/cvpr/GeigerLU12,DBLP:journals/pami/DollarWSP12,DBLP:journals/ijcv/EveringhamEGWWZ15,DBLP:conf/eccv/LinMBHPRDZ14} 
constructed to promote the developments in related fields. PASCAL VOC \cite{DBLP:journals/ijcv/EveringhamEGWWZ15} is one of the most popular benchmarks in generic object detection including $20$ classes, \eg, \textit{person}, \textit{aeroplane}, \textit{car} and \textit{tvmonitor}. Recently, MS COCO \cite{DBLP:conf/eccv/LinMBHPRDZ14} with $80$ common object categories in real life, becomes the mainstream benchmark for object detection. It is more challenging than PASCAL VOC \cite{DBLP:journals/ijcv/EveringhamEGWWZ15} due to several factors such as scale variations of objects, clutter background and heavy occlusion.

{\noindent {\bf Video object detection datasets.}}
To advance the development of object detection in videos, ILSVRC \cite{DBLP:journals/ijcv/RussakovskyDSKS15} selects $30$ categories of real-life objects for video object detection. The YouTube-Objects dataset \cite{DBLP:conf/cvpr/PrestLCSF12} is another large-scale dataset collected from YouTube, which is formed by $9\sim24$ annotated videos in $10$ categories. Meanwhile, Wen \etal \cite{DBLP:journals/corr/WenDCLCQLYL15,DBLP:conf/avss/LyuCDWQLWKHCCAB17,DBLP:conf/avss/LyuCDLWCCSMDCBG18} collect the UA-DETRAC dataset in traffic scenarios for vehicle detection. The MOTChallenge team constructs the MOT17Det \cite{DBLP:journals/ijcv/DendorferOMSCRR21} and MOT20Det \cite{DBLP:journals/corr/abs-2003-09003} benchmarks for pedestrian detection.

{\noindent {\bf Single object tracking datasets.}}
Various datasets \cite{DBLP:journals/pami/SmeuldersCCCDS14,DBLP:journals/tip/LiangBL15,DBLP:conf/iccv/GaloogahiFHRL17,DBLP:conf/icra/LiangWLWLL18,DBLP:journals/pami/WuLY15,DBLP:conf/eccv/KristanLMFPCVHL16,DBLP:conf/iccvw/KristanLMFPZVHL17,DBLP:conf/eccv/KristanLMFPZVBL18,DBLP:journals/tip/DuQLWHL16} are collected to facilitate the training and testing of single object tracking methods.
Among them, OTB \cite{DBLP:journals/pami/WuLY15} is the precursor benchmark dataset designed to fairly evaluate the single object trackers. After that, VOT \cite{DBLP:conf/eccv/KristanLMFPCVHL16,DBLP:conf/iccvw/KristanLMFPZVHL17,DBLP:conf/eccv/KristanLMFPZVBL18} becomes another popular benchmark for SOT evaluation, which includes rotated bounding box and visual attributes annotations in each frame. To evaluate the tracking accuracy of deformable objects, Du \etal \cite{DBLP:journals/tip/DuQLWHL16} propose the Deform-SOT dataset with $50$ sequences in unconstrained environments.

To facilitate data-driven deep learning methods, more large-scale visual tracking datasets have been proposed in recent three years. GOT-10k \cite{DBLP:journals/corr/abs-1810-11981} is collected by $563$ classes of moving objects and $87$ classes of motion to cover as many real-world scenarios as possible. Moreover, TrackingNet \cite{DBLP:conf/eccv/MullerBGAG18} includes more than $30K$ videos and more that $14$ million bounding box annotations to cover a wide selection of object classes in broad and diverse context. The LaSOT \cite{DBLP:journals/corr/abs-1809-07845} dataset provides a dedicated platform for training deep trackers and evaluating long-term tracking performance, to address the problems of small-scale, lack of high-quality dense annotations, short-term tracking, and category bias. 

{\noindent {\bf Multi-object tracking datasets.}}
Several multi-object tracking datasets are constructed to evaluate the MOT methods, including KITTI \cite{DBLP:conf/cvpr/GeigerLU12}, MOTChallenge \cite{DBLP:journals/ijcv/DendorferOMSCRR21}, and UA-DETRAC \cite{DBLP:journals/corr/WenDCLCQLYL15,DBLP:conf/avss/LyuCDWQLWKHCCAB17,DBLP:conf/avss/LyuCDLWCCSMDCBG18}. KITTI \cite{DBLP:conf/cvpr/GeigerLU12} is designed for autonomous driving scenarios. The MOTChallenge team establishes a unified platform to evaluate the multi-pedestrian tracking methods, including the MOT15 \cite{DBLP:journals/ijcv/DendorferOMSCRR21} and MOT16 \cite{DBLP:journals/ijcv/DendorferOMSCRR21} datasets. The UA-DETRAC dataset \cite{DBLP:journals/corr/WenDCLCQLYL15,DBLP:conf/avss/LyuCDWQLWKHCCAB17,DBLP:conf/avss/LyuCDLWCCSMDCBG18} is collected in traffic scenarios for multi-vehicle tracking. Recently, TAO \cite{DBLP:conf/eccv/DaveKTSR20} features $2,907$ HD videos in diverse environments, which is significantly larger, longer, and more diverse than the previous proposed datasets.

\subsection{Drone-based Datasets}
Besides the aforementioned object detection and tracking datasets, various drone-captured datasets are proposed in recent years for object detection \cite{DBLP:conf/iccv/HsiehLH17,DBLP:conf/cvpr/XiaBDZBLDPZ18,DBLP:conf/cvpr/BarekatainMSMNM17,DBLP:conf/fgr/KalraSN0VS19,DBLP:conf/mm/MandalKV20}, tracking \cite{DBLP:conf/eccv/MuellerSG16,DBLP:conf/eccv/RobicquetSAS16,DBLP:conf/aaai/LiY17,DBLP:conf/eccv/DuQYYDLZHT18,DBLP:journals/ijcv/YuLZHDTS20}, and semantic segmentation \cite{DBLP:journals/corr/abs-1810-10438}.

Hsieh \etal \cite{DBLP:conf/iccv/HsiehLH17} is the first aerial dataset for car counting in parking lot scenarios. After that, Xia \etal \cite{DBLP:conf/cvpr/XiaBDZBLDPZ18} propose a large-scale dataset in aerial images collected by different sensors and platforms to advance object detection research in earth vision. Barekatain \etal \cite{DBLP:conf/cvpr/BarekatainMSMNM17} develop the Okutama-Action dataset for concurrent human action detection, including $43$ minute-long fully-annotated sequences with $12$ action classes. Recently, MOR-UAV \cite{DBLP:conf/mm/MandalKV20} proposes a large-scale video dataset for the moving object recognition task, \ie, localizing and classifying the moving objects simultaneously in video frames.
 
Meanwhile, the UAV123 dataset \cite{DBLP:conf/eccv/MuellerSG16} is collected for single object tracking, captured by drones at low-altitude aerial perspectives. Li and Yeung \cite{DBLP:conf/aaai/LiY17} collect $70$ video sequences of high diversity by drone-equipped cameras for single object tracking evaluation. Robicquet \etal \cite{DBLP:conf/eccv/RobicquetSAS16} collect several video sequences using   drone-equipped cameras in campuses, including various types of objects, (\ie, pedestrians, bikes, skateboarders, cars, buses, and golf carts). Du \etal \cite{DBLP:conf/eccv/DuQYYDLZHT18,DBLP:journals/ijcv/YuLZHDTS20} also construct a UAV benchmark for three tasks, \ie, object detection, single object tracking, and multiple object tracking. Besides, the Semantic Drone dataset\footnote{\url{http://dronedataset.icg.tugraz.at/}} focuses on semantic understanding of urban scenes from bird view, including $400$ images for training and $200$ images for testing with the resolution of $6000\times4000$. Similarly, the UAVid dataset \cite{DBLP:journals/corr/abs-1810-10438} is also a UAV dataset for semantic segmentation, including $30$ video sequences in slanted views.

In contrast to the aforementioned datasets captured in limited scenarios, our {\bf \VIS} dataset is collected in various urban scenes, focusing on real world problems with new challenges, \eg, large scale and viewpoint variations, and heavy occlusions.

\section{\VIS~Overview}
A critical basis for effective algorithm evaluation is a comprehensive dataset. For this purpose, in {\bf \VIS}, we systematically collect the largest dataset to advance the object detection and tracking research on drones to date. It consists of $263$ video clips with $179,264$ frames and additional $10,209$ static images. The videos/images are acquired by various drone platforms, \ie, DJI Mavic, Phantom series (3, 3A, 3SE, 3P, 4, 4A, 4P), including different scenarios across $14$ different cites in China, \ie, Tianjin, Hongkong, Daqing, Ganzhou, Guangzhou, Jincang, Liuzhou, Nanjing, Shaoxing, Shenyang, Nanyang, Zhangjiakou, Suzhou and Xuzhou. The dataset covers various weather and lighting conditions, representing diverse scenarios in our daily life. The maximal resolutions of video clips and static images are $3840\times2160$ and $2000\times1500$, respectively.

The \VIS~benchmark focuses on the following four tasks (see Fig. \ref{fig:visdrone_example}), \ie, DET, VID, SOT and MOT. We construct a website: 
\begin{center}
\url{http://www.aiskyeye.com/}
\end{center}
for accessing the {\bf \VIS} dataset and perform evaluation of those four tasks. Notably, for each task, the images/videos in the {\tt training}, {\tt validation}, and {\tt testing} subsets are captured at different locations, but share similar scenarios and attributes. The {\tt training} subset is used to train the algorithms, the {\tt validation} subset is used to validate the performance of algorithms, the {\tt test-challenge} subset is used for workshop competition, and the {\tt test-dev} subset is used as the default test set for public evaluation. We manually annotate the bounding boxes of different categories of objects in each image or frame. After that, cross-checking is conducted to ensure annotation quality. The annotated ground-truths for {\tt training} and {\tt validation} subsets are made available to participants, but the ground-truths of the {\tt testing} subset are reserved in order to avoid (over)fitting of algorithms.

To participate our challenge, research teams are required to create their own accounts using email addresses. After registration, participants can choose the tasks of interest, and submit the results specifying locations or trajectories of objects in the images or videos using the corresponding accounts. The participants are encouraged to use the provided training data, but additional training data is allowable after declaration in submission. In the following subsections, we describe the data statistics and annotation of the datasets for each track in detail.

\section{DET Track}
The {\em DET} track tackles the problem of localizing multiple object categories in the image. For each image, algorithms are required to predict the bounding boxes of all the object instances of predefined object categories, with a real-valued confidence.

\begin{figure*}
\centering
\includegraphics[width=1.0\linewidth]{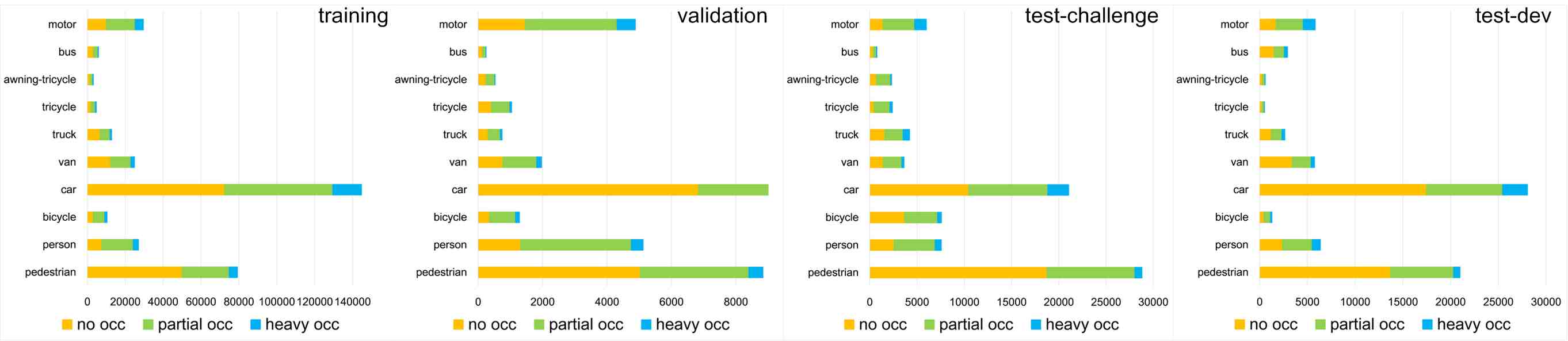}
\caption{The number of objects with different occlusion degrees of different object categories in the subsets of the {\em DET} track.}
\label{fig:image_annotation_statistics}
\end{figure*}

\subsection{Data Collection and Annotation}
The DET dataset consists of $10,209$ images in unconstrained challenging scenes, including $6,471$ images in the {\tt training} subset, $548$ in the {\tt validation} subset, $1,580$ in the {\tt test-challenge} subset, and $1,610$ in the {\tt test-dev} subset. We plot the number of objects in different object categories with different occlusion degrees in Fig. \ref{fig:image_annotation_statistics}. Notably, the class imbalance issue significantly affect the detection performance. For example, the number of the {\em awning-tricycle} instances is more than $40\times$ less than the {\em car} instances.

In this track, we mainly focus on people and vehicles in daily life and define ten object categories, including {\em pedestrian}, {\em person}\footnote{If a human maintains standing pose or walking, we classify it as {\em pedestrian}; otherwise, it is classified as a {\em person}.}, {\em car}, {\em van}, {\em bus}, {\em truck}, {\em motor}, {\em bicycle}, {\em awning-tricycle}, and {\em tricycle}. Some rarely occurring vehicles are ignored, \eg, machineshop truck, forklift truck, and tanker. We also provide two attributes of each annotated bounding box to analyze the algorithms thoroughly, \ie, the \textit{occlusion} and \textit{truncation} ratios. Specifically, occlusion ratio $\alpha$ denotes the fraction of objects being occluded by other objects or background, including no occlusion ($\delta=0\%$), partial occlusion ($\delta\in(0\%,50\%]$), and heavy occlusion ($\delta\in(50\%,100\%]$). Truncation ratio denotes the degree of object parts appearing outside a frame when the object is captured near the frame boundary. It is estimated based on the region outside the frame by human. If the truncation ratio is larger than $50\%$, the instance is not considered in evaluation.

\subsection{Evaluation Protocol}
Similar to the metrics in MS COCO \cite{DBLP:conf/eccv/LinMBHPRDZ14}, the performance of algorithms is evaluated by the {\em average precision} (AP) across different object categories and intersection over union (IoU) thresholds. Specifically, AP is computed by averaging over all $10$ IoU thresholds, \ie, in the range $[0.50:0.95]$ with uniform step size $0.05$ of all categories. AP$_{50}$ and AP$_{75}$ are computed at the single IoU thresholds $0.5$ and $0.75$ respectively. Besides, the AR$_1$, AR$_{10}$, AR$_{100}$, and AR$_{500}$ scores are the maximum recalls given $1$, $10$, $100$ and $500$ detections per image. Please refer to \cite{DBLP:conf/eccv/LinMBHPRDZ14} for more details.

\subsection{Review of Image Object Detection Methods}
Deep convolutional network dominates the object detection field in recent years, such as Faster R-CNN \cite{DBLP:journals/pami/RenHG017}, Mask R-CNN \cite{DBLP:conf/iccv/HeGDG17}, Cascade R-CNN \cite{DBLP:conf/cvpr/CaiV18}, YOLOv3 \cite{DBLP:journals/corr/abs-1804-02767}, and CenterNet \cite{DBLP:journals/corr/abs-1904-07850}. We briefly summarize some popular detection methods used in the submissions of the \VIS~workshops, which are roughly grouped into two categories, \ie, anchor-based methods and anchor-free methods. In addition, we also introduce some representative object detection methods for beginners to get started in this domain.

{\noindent \textbf{Anchor-based methods}}
predict a bounding box location and a category label for each instance in an image relying on anchor boxes\footnote{Anchor boxes are a set of pre-defined bounding boxes with various scales and aspect ratios, which are used for predicting the locations and sizes of object instances.}. It can divided into two categories, \ie, the two-stage approach, including \cite{DBLP:journals/pami/RenHG017,DBLP:conf/cvpr/LinDGHHB17,DBLP:conf/cvpr/CaiV18}, and the one-stage approach, including \cite{DBLP:conf/eccv/LiuAESRFB16,DBLP:conf/iccv/LinGGHD17,DBLP:conf/cvpr/ZhangWBLL18,DBLP:journals/corr/abs-1804-02767}. The two-stage approach generally consists of two modules. The first module is designed to generate a sparse set of object proposals and the second module is used to predict the accurate object regions and the corresponding category labels. The representative two-stage methods are described as follows.  
\begin{itemize}
\item {\bf Faster R-CNN} \cite{DBLP:journals/pami/RenHG017} is formed by the region proposal network (RPN) to first select the candidate bounding boxes of objects and Fast R-CNN \cite{DBLP:conf/iccv/Girshick15} to further generate the accurate object regions and class labels.
\item {\bf FPN} \cite{DBLP:conf/cvpr/LinDGHHB17} uses feature pyramids to build multi-scale feature maps to improve the detection accuracy.
\item {\bf Light-RCNN} \cite{DBLP:journals/corr/abs-1711-07264} relies on the thin base feature map and cheap R-CNN subnet formed by a pooling and fully-connected layers to construct high efficient detector heads. 
\item {\bf Cascade R-CNN} \cite{DBLP:conf/cvpr/CaiV18} designs the cascade architecture with a sequence of detection heads trained with the increasing IoU thresholds, to be sequentially more selective against close false positives. 
\end{itemize}

In contrast to the two-stage approach, the one-stage approach directly predicts objects by sampling over regular and dense locations, scales and aspect ratios, instead of using the time-consuming region proposal generation module. The representative methods are presented as follows. 
\begin{itemize}
\item {\bf YOLOv3} \cite{DBLP:journals/corr/abs-1804-02767} is improved from YOLO9000\cite{DBLP:conf/cvpr/RedmonF17} by making a bunch of design changes, such as designing the predictions across scales and using the feature extractor Darknet-53. After that, several YOLO successors \cite{DBLP:journals/corr/abs-2004-10934,DBLP:journals/corr/abs-2011-08036} further improve the detection accuracy using a bag of tricks. 
\item {\bf SSD} \cite{DBLP:conf/eccv/LiuAESRFB16} directly predicts the locations and scales of objects in different scales of feature maps based on a set of preset anchor boxes.
\item {\bf RetinaNet} \cite{DBLP:conf/iccv/LinGGHD17} designs the focal loss to address the class imbalance issue by reshaping the standard cross entropy loss, \ie, down-weights the loss assigned to the well-classified examples.
\item {\bf RefineDet} \cite{DBLP:conf/cvpr/ZhangWBLL18} uses the anchor refinement module to filter out negative anchors and coarsely adjust positive anchors. Then the object detection module is introduced to take the refined anchors as the input to further regress object locations and sizes and predict multi-class labels. It performs better than the two-stage methods and maintains comparable efficiency of one-stage methods.
\end{itemize}

{\noindent \textbf{Anchor-free methods}} rely on points to predict the objects instead of anchor boxes, such as corner points, center points, and keypoints, which solves the two limitations of the anchor-based methods, \ie, (1) the hand-crafted anchor boxes with predefined scales and aspect ratios are sensitive to specific dataset, and (2) densely placed anchor boxes bring huge computational cost and memory requirements. There are two anchor-free methods widely used in the challenges. That is, Duan \etal \cite{DBLP:conf/iccv/DuanBXQH019} design the CornerNet, which use the center, the top-left corner and the bottom-right corner of bounding boxes to detect each object as a triplet, improving both precision and recall. Another method is CenterNet \cite{DBLP:journals/corr/abs-1904-07850}, which models the object as a single point, \ie, the center point of its bounding box, and regresses other properties, such as object size, dimension, orientation, and pose, directly from image features at the center location. 

Meanwhile, {\bf FSAF} \cite{DBLP:journals/corr/abs-1903-00621} attaches the anchor-free branch to each level of the feature pyramid based on anchor-based detectors \cite{DBLP:conf/iccv/LinGGHD17}, allowing box encoding and decoding in the anchor-free manner at an arbitrary level. {\bf FCOS} \cite{DBLP:journals/corr/abs-1904-01355} attempts to avoid all computational cost and hyper-parameters regarding to anchor boxes and solves the object detection in the per-pixel prediction fashion, with low complexity. {\bf RepPoints} \cite{DBLP:conf/iccv/YangLHWL19} represents objects as a set of sample points critical for both localization and recognition. After that, RepPoints v2 \cite{DBLP:conf/nips/Chen00WL020} enhances the original regression-based RepPoints by integrating the verification task for better performance.

{\noindent \textbf{Other strategies.}} Besides, there are several other strategies designed to further improve the object detection accuracy.
\begin{itemize}
\item {\bf Ensemble learning} combines the predictions of multiple aforementioned models (\eg, Faster R-CNN \cite{DBLP:journals/pami/RenHG017}, Cascade R-CNN \cite{DBLP:conf/cvpr/CaiV18}, YOLOv3 \cite{DBLP:journals/corr/abs-1804-02767}) to reduce the variance of predictions and improve the performance. Besides, fusing different backbone networks\footnote{The backbone network denotes the feature extracting network, which is used to extract the features of the input images or videos.} is another alternative. For example, CBNet \cite{DBLP:conf/aaai/LiuWWLZTL20} assembles multiple identical backbones to construct a more powerful backbone by composite connections between the adjacent backbones.

\item {\bf Feature extraction.} Appearance plays a critical role in object detection. Researchers rack their brains to design different networks to exploit discriminative features to improve the performance. Inspired by FPN \cite{DBLP:conf/cvpr/LinDGHHB17}, Libra R-CNN \cite{DBLP:conf/cvpr/PangCSFOL19} strengthens multi-level features using the same deeply integrated balanced semantic features. HRNet \cite{DBLP:conf/cvpr/0009XLW19} connects the convolution layers from high to low resolutions in parallel to generate discriminative high-resolution representations.  DetNet \cite{DBLP:conf/eccv/LiPYZDS18} introduces extra stages into traditional backbone networks, while maintains high spatial resolution in deeper layers. DetectoRS \cite{DBLP:journals/corr/abs-2006-02334} develops recursive feature pyramid at macro levels to exploit extra feedback connections from FPN into the bottom-up backbone layers, and switchable atrous convolution at micro levels to convolve the features with different atrous rates. Notably, other effective modules are also introduced to further enhance the feature representations, such as spatial group-wise enhance \cite{DBLP:journals/corr/abs-1905-09646}, global context \cite{DBLP:journals/corr/abs-1904-11492}, deformable convolution \cite{DBLP:conf/iccv/DaiQXLZHW17}, squeeze-and-excitation \cite{DBLP:conf/cvpr/HuSS18}, and CARAFE \cite{DBLP:conf/iccv/WangCX0LL19}.

\item {\bf Anchor box} is important for the anchor-based object detectors. Guided anchor \cite{DBLP:journals/corr/abs-1901-03278} predicts the locations and shapes of anchor boxes by leveraging semantic features to guide the anchor box design. ATSS \cite{DBLP:conf/cvpr/ZhangCYLL20} and PAA \cite{DBLP:conf/eccv/KimL20} automatically select positive and negative anchor box samples according to the statistical characteristics of objects and the probability distributions of anchors respectively.

\item {\bf Multi-scale training} is an effective way to improve the detection accuracy. To accelerate multi-scale training, SNIPER \cite{DBLP:conf/nips/SinghND18} samples chips from multiple scales of the image pyramid in the training stage, conditioned on the image content. Since the detector is trained with the clips of $512\times512$ dimensions, it can reap the benefits of a large batch size and be trained with the batch normalization layers on a single GPU. Meanwhile, YOLOv4 \cite{DBLP:journals/corr/abs-2004-10934} develops the Mosaic data augmentation to mix $4$ training images and allows detection of objects outside their normal context, which significantly reduces the need of large mini-batch size.

\item {\bf Segmentation branch} provides more accurate feature representations with strong semantic cue for object detection. Several methods such as Mask R-CNN \cite{DBLP:conf/iccv/HeGDG17}, DeepLab \cite{DBLP:conf/cvpr/YuWPGYS18} and DES \cite{DBLP:conf/cvpr/ZhangQX0WY18} introduce the segmentation branch to predict the masks in parallel with the localizations and sizes of objects to improve the accuracy.

\item {\bf Region search} strategy demonstrates promising results in handling objects with small scales. To improve performance in cluster regions, several approaches such as ClusDet \cite{DBLP:conf/iccv/YangFCBL19} and AutoFocus \cite{DBLP:conf/iccv/NajibiS019} propose to segment the local regions for better detection of small objects.
\end{itemize} 

\begin{table}[t]
\centering
\caption{Comparison results (\ie, percentage of AP scores) of the algorithms on the \VIS-DET dataset.}
\setlength{\tabcolsep}{2pt}
\tiny{
\begin{tabular}{|c|c|ccc|cccc|}
\hline
Method   &Representative Model &AP &AP$_{50}$ &AP$_{75}$ &AR$_{1}$ &AR$_{10}$ &AR$_{100}$ &AR$_{500}$ \\
\hline
		    \multicolumn{9}{c}{\VIS-2018 challenge:}\\			
\hline
HAL-Retina-Net   &     RetinaNet \cite{DBLP:conf/iccv/LinGGHD17}    &\textbf{31.88} &46.18 &\textbf{32.12} &0.97 &7.50 &34.43 &\textbf{90.63} \\
DPNet            &     Faster R-CNN \cite{DBLP:journals/pami/RenHG017}    &30.92 &\textbf{54.62} &31.17 &1.05 &8.00 &\textbf{36.80} &50.48 \\
DE-FPN           &     FPN \cite{DBLP:conf/cvpr/LinDGHHB17}          &27.10 &48.72 &26.58 &0.90 &6.97 &33.58 &40.57 \\
CFE-SSDv2        &     SSD \cite{DBLP:conf/eccv/LiuAESRFB16}       &26.48 &47.30 &26.08 &1.16 &\textbf{8.76} &33.85 &38.94 \\
RD$^4$MS         &     RefineDet \cite{DBLP:conf/cvpr/ZhangWBLL18} &22.68 &44.85 &20.24 &1.55 &7.45 &29.63 &38.59 \\
L-H RCNN+        &     Light-RCNN \cite{DBLP:journals/corr/abs-1711-07264}     &21.34 &40.28 &20.42 &1.08 &7.81 &28.56 &35.41 \\
Faster R-CNN2    &     Faster R-CNN \cite{DBLP:journals/pami/RenHG017}    &21.34 &40.18 &20.31 &\textbf{1.36} &7.47 &28.86 &37.97 \\
RefineDet+       &     RefineDet \cite{DBLP:conf/cvpr/ZhangWBLL18} &21.07 &40.98 &19.65 &0.78 &6.87 &28.25 &35.58 \\
DDFPN            &     FPN \cite{DBLP:conf/cvpr/LinDGHHB17}       &21.05 &42.39 &18.70 &0.60 &5.67 &28.73 &36.41 \\
YOLOv3\_DP       &     YOLOv3 \cite{DBLP:journals/corr/abs-1804-02767}      &20.03 &44.09 &15.77 &0.72 &6.18 &26.53 &33.27 \\
\hline
		    \multicolumn{9}{c}{\VIS-2019 challenge:}\\			
\hline
    DPNet-ensemble &Cascade R-CNN \cite{DBLP:conf/cvpr/CaiV18}& \textbf{29.62}  & \textbf{54.00}  & \textbf{28.70}  & 0.58  & 3.69  & 17.10  & 42.37  \\
    RRNet &Zhou \etal's CenterNet \cite{DBLP:journals/corr/abs-1904-07850}& 29.13  & 55.82  & 27.23  & \textbf{1.02}  & \textbf{8.50}  & \textbf{35.19}  & \textbf{46.05}  \\
    ACM-OD &FPN \cite{DBLP:conf/cvpr/LinDGHHB17}& 29.13  & 54.07  & 27.38  & 0.32  & 1.48  & 9.46  & 44.53  \\
    S+D   &Cascade R-CNN \cite{DBLP:conf/cvpr/CaiV18}& 28.59  & 50.97  & 28.29  & 0.50  & 3.38  & 15.95  & 42.72  \\
    BetterFPN &FPN \cite{DBLP:conf/cvpr/LinDGHHB17}& 28.55  & 53.63  & 26.68  & 0.86  & 7.56  & 33.81  & 44.02  \\
    HRDet+ &HRDet \cite{DBLP:conf/cvpr/0009XLW19}& 28.39  & 54.53  & 26.06  & 0.11  & 0.94  & 12.95  & 43.34  \\
    CN-DhVaSa &Zhou \etal's CenterNet \cite{DBLP:journals/corr/abs-1904-07850}& 27.83  & 50.73  & 26.77  & 0.00  & 0.18  & 7.78  & 46.81  \\
    SGE-cascade R-CNN &Cascade R-CNN \cite{DBLP:conf/cvpr/CaiV18}& 27.33  & 49.56  & 26.55  & 0.48  & 3.19  & 11.01  & 45.23  \\
    EHR-RetinaNet &RetinaNet \cite{DBLP:conf/iccv/LinGGHD17}& 26.46  & 48.34  & 25.38  & 0.87  & 7.87  & 32.06  & 38.42  \\
    CNAnet &Cascade R-CNN \cite{DBLP:conf/cvpr/CaiV18}& 26.35  & 47.98  & 25.45  & 0.94  & 7.69  & 32.98  & 42.28  \\
\hline
		    \multicolumn{9}{c}{\VIS-2020 challenge:}\\			
\hline
    DroneEye2020 &Cascade R-CNN \cite{DBLP:conf/cvpr/CaiV18}&\bf{34.57} &58.21 &\bf{35.74} &0.28 &1.92 &6.93 &52.37\\
    TAUN  &ATSS \cite{DBLP:conf/cvpr/ZhangCYLL20} &34.54 &\bf{59.42} &34.97 &0.14 &0.72 &12.81 &49.80   \\
    CDNet  &Cascade R-CNN \cite{DBLP:conf/cvpr/CaiV18}&34.19 &57.52 &35.13 &0.80 &8.12 &39.39 &\bf{52.62} \\
    CascadeAdapt &Cascade R-CNN \cite{DBLP:conf/cvpr/CaiV18}&34.16 &58.42 &34.50 &0.84 &\bf{8.17} &\bf{39.96} &47.86 \\
    HR-Cascade++  &Cascade R-CNN \cite{DBLP:conf/cvpr/CaiV18}&32.47 &55.06 &33.34 &\bf{0.94} &7.81 &37.93 &50.65  \\       
    MSC-CenterNet &Zhou \etal's CenterNet \cite{DBLP:journals/corr/abs-1904-07850}&31.13 &54.13 &31.41 &0.27 &1.85 &6.12 &50.48  \\
    CenterNet+  &Zhou \etal's CenterNet \cite{DBLP:journals/corr/abs-1904-07850}&30.94 &52.82 &31.13 &0.27 &1.84 &5.67 &50.93 \\
    ASNet  &ATSS \cite{DBLP:conf/cvpr/ZhangCYLL20} &29.57 &52.25 &29.37 &0.25 &1.69 &6.46 &46.01  \\
    CN-FaDhSa  &Zhou \etal's CenterNet \cite{DBLP:journals/corr/abs-1904-07850}&28.52 &49.50 &28.86 &0.26 &1.76 &6.32 &48.06  \\
    HRNet  &Duan \etal's CenterNet \cite{DBLP:conf/iccv/DuanBXQH019}&27.39 &49.90 &26.71 &0.80 &7.67 &33.67 &46.16 \\  
\hline	
		    \multicolumn{9}{c}{\VIS-dev:}\\	
\hline	
    CornerNet \cite{DBLP:conf/eccv/LawD18} &- & \textbf{23.43} & \textbf{41.18} & \textbf{25.02} & \textbf{0.45} & \textbf{4.24} & \textbf{33.05} & \textbf{34.23} \\
    Light-RCNN \cite{DBLP:journals/corr/abs-1711-07264} &- & 22.08 & 39.56 & 23.24 & 0.32 & 3.63 & 31.19 & 32.06 \\
    FPN \cite{DBLP:conf/cvpr/LinDGHHB17}   &- & 22.06 & 39.57 & 22.50 & 0.29 & 3.50 & 30.64 & 31.61 \\
    Cascade R-CNN \cite{DBLP:conf/cvpr/CaiV18} &- & 21.80 & 37.84 & 22.56 & 0.28 & 3.55 & 29.15 & 30.09 \\
    DetNet \cite{DBLP:conf/eccv/LiPYZDS18} &- & 20.07 & 37.54 & 21.26 & 0.26 & 2.84 & 29.06 & 30.45 \\
    RefineDet \cite{DBLP:conf/cvpr/ZhangWBLL18} &- & 19.89 & 37.27 & 20.18 & 0.24 & 2.76 & 28.82 & 29.41 \\
    RetinaNet \cite{DBLP:conf/iccv/LinGGHD17} &- & 18.94 & 31.67 & 20.25 & 0.14 & 0.68 & 7.31 & 27.59 \\  		    	
\hline
\end{tabular}}
\label{tab:det-res}
\end{table}

\subsection{Results and Analysis}
{\flushleft {\bf Results on the {\tt test-challenge} set}.}
Top $10$ object detectors in the \VIS-DET2018 \cite{DBLP:conf/eccv/ZhuWDBLHNCLLMWW18}, \VIS-DET2019 \cite{DBLP:conf/iccv/DuZWBLHPDET19} and \VIS-DET2020 \cite{DBLP:conf/eccv/DuWZFHLSPASPJDL20} challenges are presented in Table \ref{tab:det-res}. In contrast to existing object detection datasets, \eg, MS COCO \cite{DBLP:conf/eccv/LinMBHPRDZ14} and UA-DETRAC \cite{DBLP:journals/corr/WenDCLCQLYL15}, one of the most challenging issues in the \VIS-DET dataset is the extremely small scale of objects.

As shown in Table \ref{tab:det-res}, we find that HAL-Retina-Net and DPNet are the only two methods achieving higher than $30\%$ AP in the \VIS-DET2018 challenge \cite{DBLP:conf/eccv/ZhuWDBLHNCLLMWW18}. Specifically, HAL-Retina-Net uses the squeeze-and-excitation \cite{DBLP:conf/cvpr/HuSS18} and downsampling-upsampling \cite{DBLP:conf/cvpr/WangJQYLZWT17} modules to learn both the channel and spatial attentions\footnote{Inspired by human visual system, attention mechanism can guide the network to grasp important contents within different channels or spatial regions of the feature maps corresponding to images/videos.} on multi-scale features. To detect small scale objects, it removes higher convolutional layers in the feature pyramid. The second best detector DPNet uses FPN \cite{DBLP:conf/cvpr/LinDGHHB17} to extract multi-scale features and uses ensemble mechanism to combine three detectors with different backbones, \ie, ResNet-50, ResNet-101 and ResNeXt. Following DPNet, DE-FPN and CFE-SSDv2 also employ multi-scale features to achieve good performance. RD$^4$MS trains $4$ variants of RefineDet \cite{DBLP:conf/cvpr/ZhangWBLL18}, \ie, three use SEResNeXt-50 and one uses ResNet-50 as the backbone network. Moreover, DDFPN introduces deep back-projection super-resolution network \cite{DBLP:conf/cvpr/HarisSU18} to upsample the image using the deformable FPN architecture \cite{DBLP:conf/iccv/DaiQXLZHW17}. Notably, most of the submitted methods use multi-scale testing strategy to improve detection performance effectively.

In the \VIS-DET2019 challenge \cite{DBLP:conf/iccv/DuZWBLHPDET19}, DPNet-ensemble achieves the best results with $29.62\%$ AP score. It uses the global context module \cite{DBLP:journals/corr/abs-1904-11492} to integrate context information and deformable convolution \cite{DBLP:conf/iccv/DaiQXLZHW17} to enhance the transformation modeling capability of the detector. RRNet and ACM-OD tie for the second place in ranking with $29.13\%$ AP score. RRNet is improved from \cite{DBLP:journals/corr/abs-1904-07850} by integrating a re-regression module, formed by the ROIAlign module \cite{DBLP:conf/iccv/HeGDG17} and several convolution layers. ACM-OD introduces an active learning strategy, which is conducted with data augmentation for better performance. Compared with \VIS-DET2018 challenge \cite{DBLP:conf/eccv/ZhuWDBLHNCLLMWW18}, the submissions in \VIS-DET2019 challenge \cite{DBLP:conf/iccv/DuZWBLHPDET19} fail to find other effective strategies to further improve the performance.

In the \VIS-DET2020 challenge \cite{DBLP:conf/eccv/DuWZFHLSPASPJDL20}, the use of the Cascade R-CNN \cite{DBLP:conf/cvpr/CaiV18} framework has become wide-spread due to its high performance and easy extensibility. Compared with the baseline Cascade R-CNN \cite{DBLP:conf/cvpr/CaiV18} with the mAP score of $16.09\%$, the submitted varaints largely improve the performance by combining several effective modules. DroneEye2020 is mainly based on Cascade R-CNN \cite{DBLP:conf/cvpr/CaiV18} with recursive feature pyramid and switchable strous convolution \cite{DBLP:journals/corr/abs-2006-02334}, achieving the best performance with $34.57$ mAP. TAUN uses mean teacher \cite{DBLP:conf/nips/TarvainenV17} to train the cascade DetectoRS model \cite{DBLP:conf/cvpr/CaiV18,DBLP:journals/corr/abs-2006-02334}, which performs similarly as DroneEye2020. CDNet and CascadeAdapt combine Cascade R-CNN \cite{DBLP:conf/cvpr/CaiV18} with deformable convolutions, and then improve the detection accuracy using several data augmentation strategies such as sub-image splitting and mosaic \cite{DBLP:journals/corr/abs-2004-10934}. These results indicate that the detection accuracy of small objects can be improved by enhancing IoU thresholds to train multiple localization branches.

{\noindent {\bf Results on the {\tt test-dev} set}.}
For the {\tt test-dev} set, CornerNet \cite{DBLP:conf/eccv/LawD18} achieves the top AP score of $23.43\%$, which uses the Hourglass-104 backbone for feature extraction. In contrast to FPN \cite{DBLP:conf/cvpr/LinDGHHB17} and RetinaNet \cite{DBLP:conf/cvpr/ZhangWBLL18} with extra stages against the image classification task to handle objects with various scales, DetNet \cite{DBLP:conf/eccv/LiPYZDS18} re-designs the backbone network for object detection, which maintains the spatial resolution and enlarges the receptive field, achieving $20.07\%$ AP score. Meanwhile, RefineDet \cite{DBLP:conf/cvpr/ZhangWBLL18} with the VGG-16 backbone performs better than RetinaNet \cite{DBLP:conf/iccv/LinGGHD17} with the ResNet-101 backbone, \ie, $19.89\%$ {\em vs.} $18.94\%$ in terms of AP score. This is because RefineDet \cite{DBLP:conf/cvpr/ZhangWBLL18} uses the object detection module to regress the locations and sizes of objects based on the coarsely adjusted anchors from the anchor refinement module.

\subsection{Discussion}
To describe the appearance of small objects, top performers share some effective feature extraction modules such as feature pyramid \cite{DBLP:conf/cvpr/LinDGHHB17}, attention mechanism \cite{DBLP:conf/cvpr/WangJQYLZWT17} and dilated convolution \cite{DBLP:journals/corr/abs-2006-02334}. In this way, the salient semantic features can be extracted with the larger receptive field. Besides, the ensemble mechanism or cascade architecture of models \cite{DBLP:conf/cvpr/CaiV18} can further improve the performance in complex scenarios.

Although the best detector DroneEye2020 sets the new state-of-the-art in 2020, the best AP score is still less than $35\%$. Notably, it still performs not well of the small-scale objects, \eg, producing less than $25\%$ mAP score in terms of \textit{person} and \textit{bicycle}, demonstrating that the community is badly in need of developing robust methods for real-world applications. There are mainly two issues worth to explore in drone captured visual data. 

\begin{figure}[t]
\centering
\includegraphics[width=.9\linewidth]{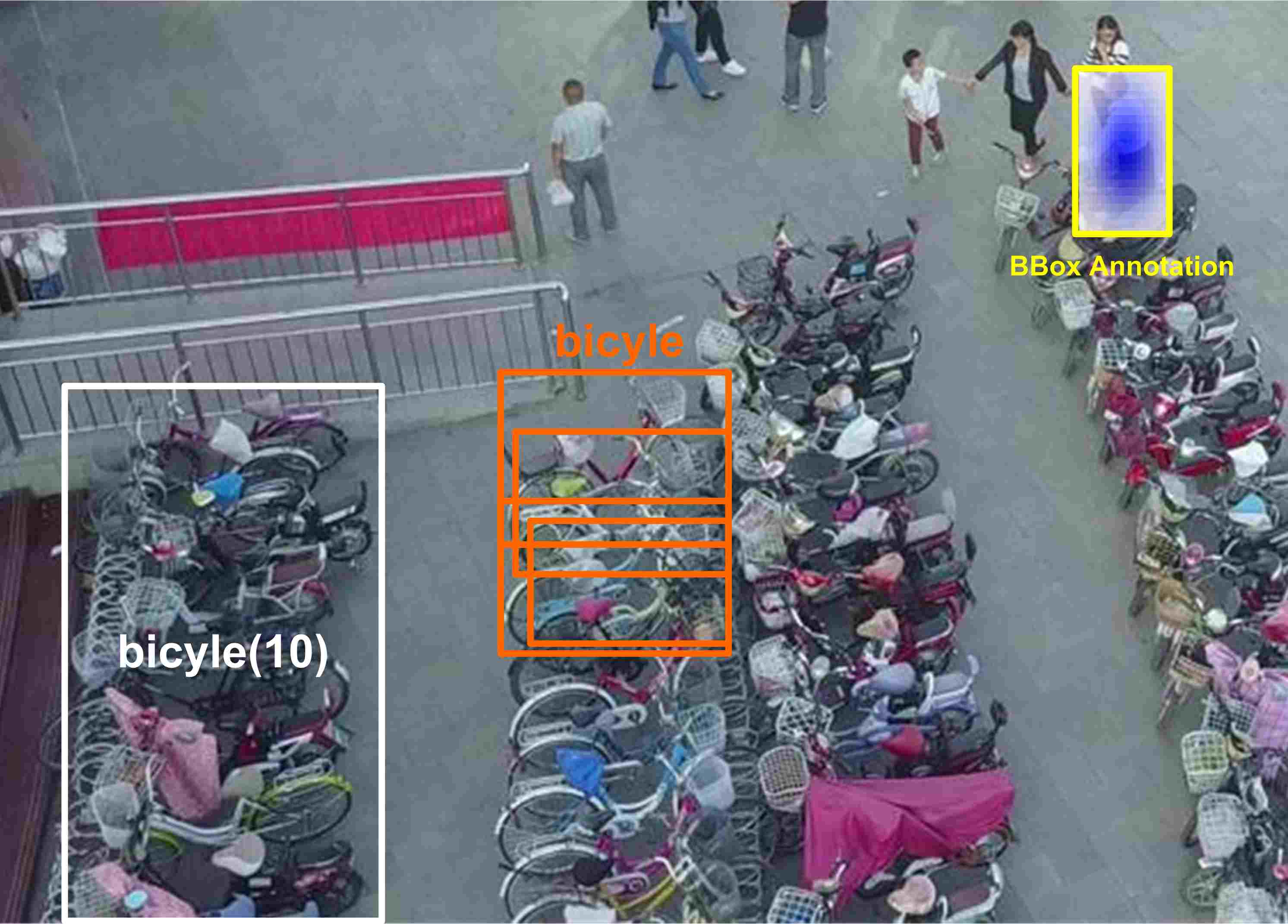}
\caption{Descriptions of the challenging issues in the image object detection task.}
\label{fig:det_anno_problem}
\end{figure}

{\noindent \textbf{Annotation and evaluation protocol}.}
As shown in Fig. \ref{fig:det_anno_problem}, there are groups of objects heavily occluded in drone captured visual data (see the orange bounding boxes of bicycles). If we use Non-maximum Suppression (NMS) to suppress duplicate detections in detectors, the majority of true positive objects will be inevitably removed. In some real applications, it is unnecessary and impractical to locate each individual object in the crowd. Thus, it is more reasonable to use a large bounding box with a count number to represent the group of objects in the same category (see the white bounding box of bicycle). Meanwhile, if we use the new annotation remedy, we need to redesign the metric to evaluate detection algorithms, \ie, both the localization and counting accuracy should be considered in evaluation \cite{DBLP:conf/aaai/CaiWZDW21}.

{\noindent \textbf{Coarse segmentation}.}
Current object detection methods use bounding boxes to indicate object instances, \ie, a $4$-tuple $(x, y, w, h)$, where $x$ and $y$ are the coordinate of the bounding box's top-left corner, and $w$ and $h$ are the width and height of the bounding box. As shown in Fig. \ref{fig:det_anno_problem}, it is difficult to predict the accurate location and size of the pedestrian (see the yellow bounding box) due to occlusion and non-rigid deformation of human body. A possible way to mitigate such issue is to integrate coarse segmentation into object detection, which might be effective to remove the disturbance of background area enclosed in the bounding box of non-rigid objects, \eg, {\em person} and {\em bicycle}, see Fig. \ref{fig:det_anno_problem}. In summary, this interesting problem is still far from being solved and worth to explore.

\section{VID Track}
The {\em VID} track aims to locate object instances from a pre-defined set of categories in the video sequences. That is, given a series of video clips, the algorithms are required to produce a set of bounding boxes of each object instance in each video frame (if any), with real-valued confidences. In contrast to {\em DET} track focusing on object detection in individual images, we deal with detecting object instances in video clips, which involve temporal consistency in consecutive frames. Five categories of objects are considered in this track, \ie, {\em pedestrian}, {\em car}, {\em van}, {\em bus}, and {\em truck}. We use the same metrics in the {\em DET} track to evaluate the video object detection algorithms.

\begin{figure}[t]
\centering
\includegraphics[width=1.0\linewidth]{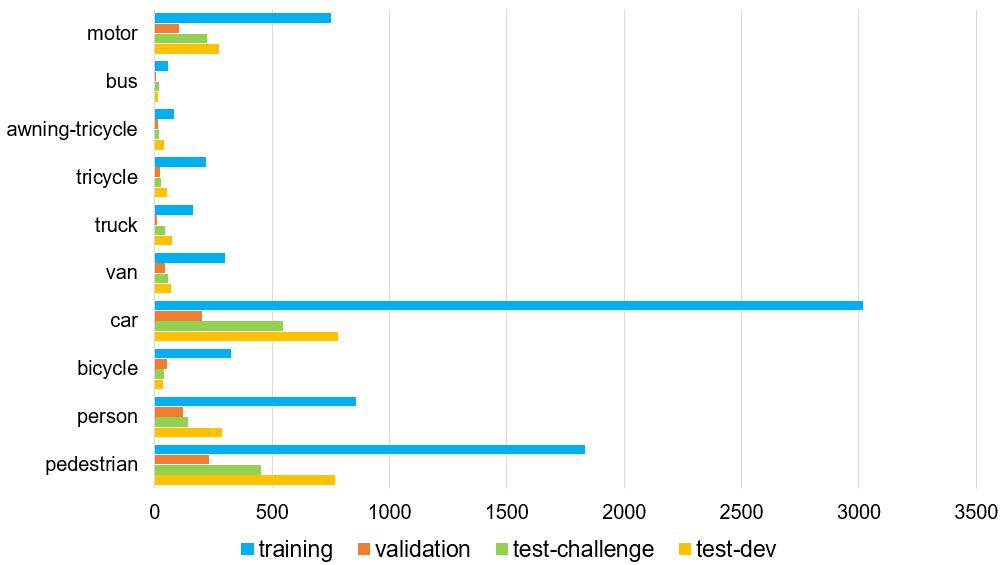}
\caption{The number of object trajectories in different categories in the subsets of the {\em VID} and {\em MOT} tracks.}
\label{fig:video_annotation_statistics}
\end{figure}

\subsection{Data Collection and Annotation}
We provide $96$ challenging video clips in the {\em VID} track, including $56$ clips for training ($24,198$ frames in total), $7$ for validation ($2,846$ frames in total), $16$ for challenge testing ($6,322$ frames in total) and $17$ for dev testing ($6,635$ frames in total). We plot the number of objects of different object categories in Fig. \ref{fig:video_annotation_statistics}. It is worth mentioning that the class imbalance issue is extremely severe in our dataset, challenging the performance of algorithms. For example, in the {\tt training} set, the number of {\em car} trajectories is more than $50\times$ of the number of {\em bus} trajectories. Meanwhile, the length of object trajectories varies dramatically, \eg, the maximal and minimal lengths of object trajectories are $1$ and $1,255$, requiring the tracking algorithms to perform well in both the short-term and long-term tracking scenarios. Similar to the {\em DET} track, we also provide the annotations of occlusion and truncation ratios of each object and ignored regions in each video frame.

\subsection{Review of Video Object Detection Methods}
Among all submissions in the challenges, the majority of them are directly derived from the image detectors such as SSD \cite{DBLP:conf/eccv/LiuAESRFB16}, Cascade R-CNN \cite{DBLP:conf/cvpr/CaiV18}, FCOS \cite{DBLP:journals/corr/abs-1904-01355}, CenterNet \cite{DBLP:journals/corr/abs-1904-07850}, and RetinaNet \cite{DBLP:conf/iccv/LinGGHD17}, and integrate the temporal coherence of video sequences to handle appearance deterioration such as motion blur, video defocus of objects. 

{\noindent \textbf{Data association}} is an effective way to exploit temporal consistency of detections in consecutive frames. A simple solution is to combine image detectors and multiple single object trackers (\eg, ECO \cite{DBLP:conf/cvpr/DanelljanBKF17} and SiamRPN++ \cite{DBLP:journals/corr/abs-1812-11703}) for video object detection. Several submissions in the \VIS ~challenges adopt this strategy to improve the accuracy, such as EODST in the \VIS-VDT2018 challenge \cite{DBLP:conf/eccv/ZhuWDBLHWNCLLMW18} and EODST+ in the VID2019 \cite{DBLP:conf/iccv/ZhuDWBLHPZWZBS19} challenge. Moreover, some other methods extend object detections in individual frames into tracks or tubelets. The representative methods are presented as follows. 
\begin{itemize}
\item \textbf{T-CNN} \cite{DBLP:journals/corr/KangLYZYXZWWWO16} uses optical flow to propagate the detected bounding box across frames and links the bounding box into tubelets with tracking algorithms. 
\item \textbf{Seq-NMS} \cite{DBLP:journals/corr/HanKPRBSLYH16} re-scores all detected bounding boxes within a video sequence to search the optimal box linkage.
\item \textbf{D\&T} \cite{DBLP:conf/iccv/FeichtenhoferPZ17} is an end-to-end trained CNN to simultaneously solve detection and tracking without box-level post-processing, which uses a multi-task loss for frame-level object detection and cross-frame track regression in training. 
\end{itemize}
However, long time partial or full occlusion, severe camera view changes, and fast motion are significant challenging factors in drone-captured videos, making data association based approach less effective.

{\noindent \textbf{Feature aggregation}} fuses object features in multiple video frames to improve performance for video object detection. The representative methods as summarized as follows.
\begin{itemize}
\item \textbf{FGFA+} in the \VIS-VDT2018 challenge \cite{DBLP:conf/eccv/ZhuWDBLHWNCLLMW18,DBLP:conf/iccv/ZhuWDYW17} leverages optical flow based temporal coherence to aggregate feature maps of nearby frames.
\item \textbf{DFF} \cite{DBLP:conf/cvpr/ZhuXDYW17} adopts in-network fine-tuned optical flow to propagate and align the features of selected keyframes to nearby non-keyframes, thus reducing redundant calculation and speeding up the system. 
\item \textbf{MANet} \cite{DBLP:conf/eccv/WangZYD18} combines instance and pixel-level feature calibration and aggregation through a motion pattern reasoning module.
\item \textbf{MEGA} \cite{DBLP:conf/cvpr/Chen0HW20} extracts keyframes to get access to more content using the long range memory module, and enhance their features with both global and local semantic information.
\end{itemize}
However, most of the aforementioned methods combine features of just a few consecutive frames, which are less effective to describe the long-term dynamics of objects.

{\noindent \textbf{Recurrent neural networks (RNNs)}} leverage rich temporal information to capture long-range temporal context information in videos.
\begin{itemize}
\item \textbf{a\_LSTM} \cite{DBLP:conf/iccv/LuLT17} regresses object locations and categories directly, which uses the association features to represent detected objects. The feature representations are optimized by minimizing the association error.
\item \textbf{STMM} \cite{DBLP:conf/eccv/XiaoL18} serves as the recurrent computation unit to model long-term temporal appearance and motion dynamics, where the spatial-temporal memory is aligned from frame to frame.
\item \textbf{OGEMN} \cite{DBLP:conf/iccv/DengHSZXMRG19} is an object guided external memory network with write and read operations to efficiently store and accurately propagate multi-level features.
\end{itemize}
Notably, RNNs suffer from the vanishing gradient problem. It is doubtful that the aforementioned methods can deal with complex scenes in the drone-captured videos that demand the long-range temporal context to help generate accurate results.

\begin{table}[t]
\centering
\caption{Comparison results (\ie, percentage of AP scores) of the algorithms on the \VIS-VID dataset.}
\setlength{\tabcolsep}{2.5pt}
\tiny{
\begin{tabular}{|c|c|ccc|cccc|}
\hline
Method   &Representative Model &AP &AP$_{50}$ &AP$_{75}$ &AR$_{1}$ &AR$_{10}$ &AR$_{100}$ &AR$_{500}$\\
\hline
		    \multicolumn{9}{c}{\VIS-2018 challenge:}\\			
\hline
CFE-SSDv2    &SSD \cite{DBLP:conf/eccv/LiuAESRFB16}                 &\textbf{21.57} &\textbf{44.75} &\textbf{17.95} &\textbf{11.85} &\textbf{30.46} &\textbf{41.89} &\textbf{44.82}\\
EODST        &SSD \cite{DBLP:conf/eccv/LiuAESRFB16}                 &16.54 &38.06 &12.03 &10.37 &22.02 &25.52 &25.53\\
FGFA+        &FGFA \cite{DBLP:conf/iccv/ZhuWDYW17}                  &16.00 &34.82 &12.65 &9.63 &19.54 &22.37 &22.37\\
RD           &RefineDet \cite{DBLP:conf/cvpr/ZhangWBLL18}                 &14.95 &35.25 &10.11 &9.67 &24.60 &29.72 &29.91\\
CERTH-ODV    &Faster R-CNN \cite{DBLP:journals/pami/RenHG017}                 &9.10 &20.35 &7.12 &7.02 &13.51 &14.36 &14.36\\
RetinaNet\_s &RetinaNet \cite{DBLP:conf/iccv/LinGGHD17}                 &8.63 &21.83 &4.98 &5.80 &12.91 &15.15 &15.15\\
\hline
		    \multicolumn{9}{c}{\VIS-2019 challenge:}\\			
\hline
    DBAI-Det &Cascade R-CNN \cite{DBLP:conf/cvpr/CaiV18}& \textbf{29.22} & \textbf{58.00} & \textbf{25.34} & \textbf{14.30} & \textbf{35.58} & \textbf{50.75} & \textbf{53.67} \\
    AFSRNet &RetinaNet \cite{DBLP:conf/iccv/LinGGHD17}& 24.77 & 52.52 & 19.38 & 12.33 & 33.14 & 45.14 & 45.69 \\
    HRDet+ &HRNet \cite{DBLP:conf/cvpr/0009XLW19} & 23.03 & 51.79 & 16.83 & 4.75  & 20.49 & 38.99 & 40.37 \\
    VCL-CRCNN &Cascade R-CNN \cite{DBLP:conf/cvpr/CaiV18}& 21.61 & 43.88 & 18.32 & 10.42 & 25.94 & 33.45 & 33.45 \\
    CN-DhVaSa &Zhou \etal's~CenterNet \cite{DBLP:journals/corr/abs-1904-07850}& 21.58 & 48.09 & 16.76 & 12.04 & 29.60 & 39.63 & 40.42 \\
    DetKITSY &Cascade R-CNN \cite{DBLP:conf/cvpr/CaiV18}& 20.43 & 46.33 & 14.82 & 8.64  & 25.80 & 33.40 & 33.40 \\
    EODST++ &SSD \cite{DBLP:conf/eccv/LiuAESRFB16}, FCOS \cite{DBLP:journals/corr/abs-1904-01355}& 18.73 & 44.38 & 12.68 & 9.67  & 22.84 & 27.62 & 27.62 \\
    Libra-HBR &Libra R-CNN \cite{DBLP:conf/cvpr/PangCSFOL19}& 18.29 & 44.92 & 11.64 & 10.69 & 26.68 & 35.83 & 36.57 \\
    Sniper+ &Faster R-CNN \cite{DBLP:journals/pami/RenHG017} & 18.16 & 38.56 & 14.79 & 9.98  & 27.18 & 38.21 & 39.08 \\
    FRFPN &Faster R-CNN \cite{DBLP:journals/pami/RenHG017}&16.50 & 40.15 & 11.39 & 9.72 &22.55 &28.40 &28.40 \\
\hline
		    \multicolumn{9}{c}{\VIS-dev:}\\			
\hline
    FGFA \cite{DBLP:conf/iccv/ZhuWDYW17}  &- & \textbf{14.44} & \textbf{33.34} & \textbf{11.85} & 7.29  & \textbf{21.37} & \textbf{27.09} & \textbf{27.21} \\
    D\&T \cite{DBLP:conf/iccv/FeichtenhoferPZ17}  &-  & 14.21 & 32.28 & 10.39 & \textbf{7.59}  & 19.39 & 26.57 & 25.64 \\
    FPN \cite{DBLP:conf/cvpr/LinDGHHB17}  &- & 12.93 & 29.88 & 10.12 & 7.03  & 19.71 & 25.59 & 25.59 \\
    CenterNet \cite{DBLP:journals/corr/abs-1904-07850} &- & 12.35 & 28.93 & 9.92  & 6.41  & 18.93 & 24.87 & 24.87 \\
    CornerNet \cite{DBLP:conf/eccv/LawD18} &- & 12.29 & 28.37 & 9.48  & 6.07  & 18.60  & 24.03 & 24.03 \\
    Faster R-CNN \cite{DBLP:journals/pami/RenHG017} &- & 10.25 & 26.83 & 6.70   & 5.93  & 12.98 & 13.55 & 13.55 \\
\hline

\end{tabular}}
\label{tab:vid-res}
\end{table}

\subsection{Results and Analysis}
{\noindent {\bf Results on the {\tt test-challenge} set}.}
We report the evaluation results of the submissions in the \VIS-VDT2018 \cite{DBLP:conf/eccv/ZhuWDBLHWNCLLMW18} and \VIS-VID2019 \cite{DBLP:conf/iccv/ZhuDWBLHPZWZBS19} challenges in Table \ref{tab:vid-res}. All the submitted methods focus on dealing with degenerated object appearances in videos by enhancing features or exploiting temporal coherence.

In the \VIS-VDT2018 challenge \cite{DBLP:conf/eccv/ZhuWDBLHWNCLLMW18}, CFE-SSDv2 produces the best AP score of $21.57\%$, which is improved from SSD \cite{DBLP:conf/eccv/LiuAESRFB16} by integrating a feature enhancement module for accurate results. Different from CFE-SSDv2, EODST exploits temporal information to associate object detections in individual frames using the ECO tracking method \cite{DBLP:conf/cvpr/DanelljanBKF17}, achieving the second best AP score of $16.54\%$. The third best FGFA+ is a variant of video object detection method FGFA \cite{DBLP:conf/iccv/ZhuWDYW17} by using several different data augmentation strategies.

Researchers propose several powerful methods improved from the state-of-the-art detectors, such as HRDet \cite{DBLP:conf/cvpr/0009XLW19}, Cascade R-CNN \cite{DBLP:conf/cvpr/CaiV18}, CenterNet \cite{DBLP:journals/corr/abs-1904-07850}, RetinaNet \cite{DBLP:conf/iccv/LinGGHD17}, in the \VIS-VID2019 challenge \cite{DBLP:conf/iccv/ZhuDWBLHPZWZBS19}. Notably, all top $5$ detectors, \ie, DBAI-Det, AFSRNet, HRDet+, VCL-CRCNN and CN-DhVaSa, surpass the top detector CFE-SSDv2 in the \VIS-VDT2018 challenge \cite{DBLP:conf/eccv/ZhuWDBLHWNCLLMW18}. DBAI-Det achieves the best results with $29.22\%$ AP, which integrates the deformable convolution \cite{DBLP:conf/iccv/DaiQXLZHW17} and global context \cite{DBLP:journals/corr/abs-1904-11492} in Cascade R-CNN \cite{DBLP:conf/cvpr/CaiV18}. AFSRNet ranks the second place with $24.77\%$ AP. It integrates the feature selected anchor-free head \cite{DBLP:journals/corr/abs-1903-00621} into RetinaNet \cite{DBLP:conf/iccv/LinGGHD17}. HRDet+, VCL-CRCNN and CN-DhVaSa are improved from HRDet \cite{DBLP:conf/cvpr/0009XLW19}, Cascade R-CNN \cite{DBLP:conf/cvpr/CaiV18}, and CenterNet \cite{DBLP:journals/corr/abs-1904-07850}, respectively. To deal with the large-scale variations of objects, some detectors, such as DetKITSY and EODST++, employ multi-scale features for detection, which performs better than the state-of-the-art detector FGFA \cite{DBLP:conf/iccv/ZhuWDYW17}. Notably, most of the aforementioned methods are computationally expensive for practical applications, \ie, the running speeds are less than $10$ fps on a workstation with one GTX 1080Ti GPU.

{\noindent {\bf Results on the {\tt test-dev} set}.}
The evaluation results of $2$ state-of-the-art video object detection methods \cite{DBLP:conf/iccv/ZhuWDYW17,DBLP:conf/iccv/FeichtenhoferPZ17}, and $4$ state-of-the-art image object detection methods \cite{DBLP:conf/cvpr/LinDGHHB17,DBLP:journals/corr/abs-1904-07850,DBLP:conf/eccv/LawD18,DBLP:journals/pami/RenHG017} on the {\tt test-dev} set are reported in Table \ref{tab:vid-res}. We find that the two video object detectors performs much better than the four image object detectors. For example, the second best video object detector D\&T \cite{DBLP:conf/iccv/FeichtenhoferPZ17} improves $1.28\%$ AP score compared to the top image object detector FPN \cite{DBLP:conf/cvpr/LinDGHHB17}, which demonstrates the importance of exploiting the temporal information in video object detection.

\subsection{Discussion}
In contrast to the DET task, VID methods suffers from the degenerated object appearances challenge in videos such as motion blur, pose variations, and video defocus. Exploiting temporal coherence and aggregating features in consecutive frames are two effective ways to handle such challenges. However, we observe that the majority of submitted methods rely on image based object detectors, resulting in inferior detection performance. A few submissions with additional tracking module are not matured in modeling temporal consistency of objects in consecutive frames. Notably, all the submitted methods produce less than $15\%$ mAP score on the \textit{person} and \textit{bicycle} categories. In summary, the way to exploit temporal information is still an open question for VID.

{\noindent \textbf{Temporal coherence}.}  
A feasible way to exploit temporal coherence is integrating object trackers, \eg, ECO \cite{DBLP:conf/cvpr/DanelljanBKF17} and SiamRPN++ \cite{DBLP:journals/corr/abs-1812-11703}, into object detection methods. For example, a tracker could be assigned to each detected object instance in individual frames to guide object detection in the following frames, which is effective to suppress false negatives of detectors. Meanwhile, integrating re-identification module into the detectors is another promising way to exploit temporal coherence, as described in D\&T \cite{DBLP:conf/iccv/FeichtenhoferPZ17}.

{\noindent \textbf{Feature aggregation}.} 
Aggregating features in consecutive frames is also a useful strategy to improve accuracy. As stated in FGFA \cite{DBLP:conf/iccv/ZhuWDYW17}, aggregating features of nearby frames along the motion paths to leverage temporal coherence significantly improves the performance. Thus, we can take several consecutive frames as input, and feed them into deep neural networks to extract temporal salient features using $3$D convolution operations or optical flow methods.

\begin{figure*}[t]
\centering
\includegraphics[width=1.0\linewidth]{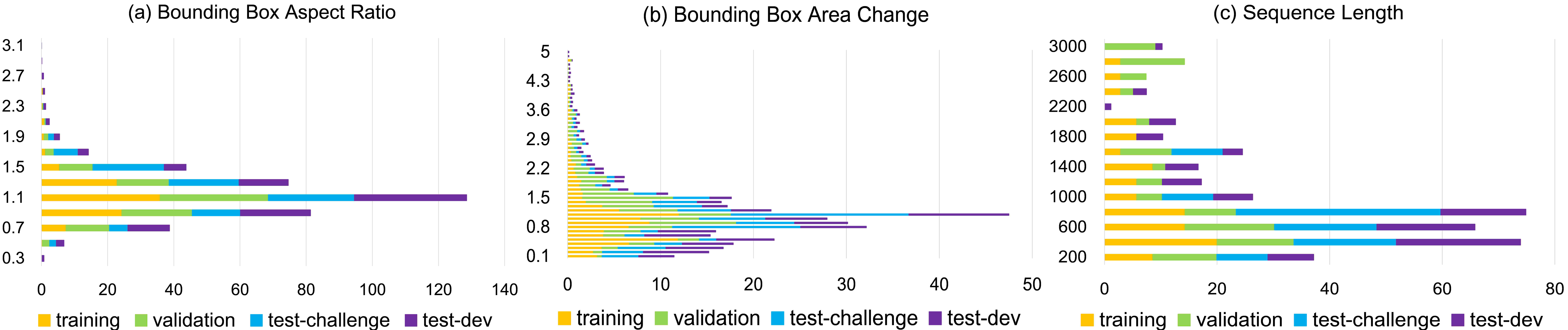}
\caption{(a) The number of frames {\em vs.} the aspect ratio (height divided by width) change rate with respect to the first frame, (b) the number of frames {\em vs.} the area variation rate with respect to the first frame, and (c) the distributions of the number of frames of video clips, in the subsets for the {\em SOT track}.}
\label{fig:sot_number_distribution}
\end{figure*}

\section{SOT Track}
For the {\em SOT} track, we focus on generic single object tracking, also known as model-free tracking \cite{DBLP:journals/pami/WuLY15}. In particular, for an input video sequence and the initial bounding box of the target object in the first frame, the {\em SOT} track requires the algorithms to locate the target bounding boxes in the subsequent video frames.

\subsection{Data Collection and Annotation}
In 2018, we provide $167$ video sequences with $139,276$ fully annotated frames, split into four subsets, \ie, the {\tt training} set ($86$ sequences with $69,941$ frames in total), {\tt validation} set ($11$ sequences with $7,046$ frames in total), {\tt test-challenge 2018} set ($35$ sequences with $29,367$ frames in total), and {\tt test-dev} set ($35$ sequences with $32,922$ frames in total). To thoroughly evaluate the performance of algorithms in long-term tracking, we add $25$ new collected sequences with $82,644$ frames in total in the {\tt test-challenge 2018} set to form the {\tt test-challenge 2019} set. The tracking targets in all these sequences include {\em pedestrian}, {\em cars}, and {\em animals}. The statistics of target objects, \ie, the aspect ratio in different frames, the area change ratio, and the sequence length are presented in Fig. \ref{fig:sot_number_distribution}.

The enclosing bounding box of target object in each video frame is annotated to evaluate the performance of trackers. To thoroughly analyze the tracking performance, we also annotate $12$ sequence attributes following \cite{DBLP:conf/eccv/MuellerSG16}, \ie, {\em aspect ratio change}, {\em background clutter}, {\em camera motion}, {\em fast motion}, {\em full occlusion}, {\em illumination variation}, {\em low resolution}, {\em out-of-view}, {\em partial occlusion}, {\em scale variation}, {\em similar object}, and {\em viewpoint change}. Please refer to \cite{DBLP:conf/eccv/MuellerSG16} for more details.

\subsection{Evaluation Protocol}
Following \cite{DBLP:journals/pami/WuLY15}, we use the success and precision scores to evaluate trackers. Specifically, we plot the success curve, \ie, the percentage of successfully tracked frames {\em vs.} bounding box overlap threshold in the range $[0,1]$. The success score is computed based on the area under the success plot, which is the primary metric for ranking trackers. The precision score denotes the percentage of frames such that the Euclidean distance between predicted locations and ground-truth locations are within $20$ pixels in the image plane. Please refer to \cite{DBLP:journals/pami/WuLY15} for more details.

\subsection{Review of Single Object Tracking Methods}
Although significant progress has been made in visual tracking, it is still difficult for the state-of-the-art trackers to produce accurate results on the drone-captured videos, due to several challenging factors such as abrupt camera motion and small scales of targets. We roughly divide them into three categories, \ie, the correlation filters based methods, the Siamese network based methods, and the convolutional network based methods. We also highlight several effective SOT methods in the challenges.

{\noindent \textbf{Correlation filters.}}
Correlation filter is one of the most popular methods in single object tracking, which produces correlation peaks for each interested target in the scene and low responses to background. It is usually implemented by using Discrete Fourier Transform, significantly reducing the storage and computational cost by several orders of magnitude. Several submissions in the \VIS-SOT2018 challenge \cite{DBLP:conf/eccv/WenZDBLHLCLMNWW18} are constructed on correlation filters, such as STALE-SRCA, DCST, LZZ-ECO, SDRCO and CFCNN. Meanwhile, we briefly review several correlation filters based methods in literature.
\begin{itemize}
\item \textbf{Staple} \cite{DBLP:conf/cvpr/BertinettoVGMT16} combines two image patch representations and constructs a model that is inherently robust to both color changes and deformations. After that, \textbf{Staple\_CA} \cite{DBLP:conf/cvpr/MuellerSG17} designs a framework that takes global context into account and incorporates it into the correlation filter trackers to boost the performance while maintaining the high frame rate. 
\item \textbf{ECO} \cite{DBLP:conf/cvpr/DanelljanBKF17} improves the discrete correlation filter in three aspects, \ie, a factorized convolution operator for parameter reduction, a generative model for better sample diversity and training efficiency, and a conservative model update strategy for robustness.
\item \textbf{C-COT} \cite{DBLP:conf/eccv/DanelljanRKF16} presents an implicit interpolation model to train multi-resolution continuous convolution filters. Meanwhile, \textbf{CFWCR} \cite{DBLP:conf/iccvw/HeFZDB17} redesigns the final confidence score function by adding the weighted sum operation, showing improvements compared to ECO \cite{DBLP:conf/cvpr/DanelljanBKF17}.
\item \textbf{BACF} \cite{DBLP:conf/iccv/GaloogahiFL17} designs a background-aware correlation filter that efficiently models the variations of both foreground and background.
\end{itemize}

{\noindent \textbf{Siamese network}.}
Besides the correlation filters, Siamese network is another popular method in the single-object tracking field with promising performance, which learns the representations of targets by minimizing the similarities of targets in consecutive frames. The representative methods include SiameseFC \cite{DBLP:conf/eccv/BertinettoVHVT16}, DSiam \cite{DBLP:conf/iccv/Guo0ZHWW17}, and SiamRPN++ \cite{DBLP:journals/corr/abs-1812-11703}. In the \VIS-SOT2019 challenge \cite{DBLP:conf/iccv/DuZWBLHZPWZBS19}, $5$ trackers use the Siamese network architecture, \ie, DC-Siam, DR-V-LT, SiamDW-FC, SiamFCOT and SiamRPN++.
\begin{itemize}
\item \textbf{SiameseFC} \cite{DBLP:conf/eccv/BertinettoVHVT16} constructs a fully-convolutional Siamese network, which is offline trained to locate an exemplar image within the search region. 
\item \textbf{DSiam} \cite{DBLP:conf/iccv/Guo0ZHWW17} is a dynamic Siamese network, which uses a fast transformation learning model to enable effective online learning of target appearance variation and background suppression from previous frames.
\item \textbf{SiamRPN++} \cite{DBLP:journals/corr/abs-1812-11703} designs a spatial-aware sampling strategy to train a ResNet-driven Siamese tracker with significant performance gain.  
\item \textbf{SiamMask} \cite{DBLP:journals/corr/abs-1812-05050} improves the offline training procedure of the fully-convolutional Siamese approaches by augmenting their loss with a binary segmentation task, \ie, producing class-agnostic object segmentation masks and rotated bounding boxes. 
\item \textbf{Siam R-CNN} \cite{DBLP:conf/cvpr/VoigtlaenderLTL20} presents a Faster R-CNN \cite{DBLP:journals/pami/RenHG017} based re-detection architecture for long-term object tracking. It designs a tracklet-based dynamic programming algorithm to take advantage of re-detections of both the first-frame template and the predictions of previous frames. 
\end{itemize}

{\noindent \textbf{CNNs}.} Some other researchers design various CNN architectures for SOT, such as MDNet \cite{DBLP:conf/cvpr/NamH16}, VITAL \cite{DBLP:journals/corr/abs-1804-04273}, CFNet \cite{DBLP:conf/cvpr/ValmadreBHVT17}, TRACA \cite{DBLP:conf/cvpr/ChoiC0YLJD018}, ATOM \cite{DBLP:journals/corr/abs-1811-07628}, and DiMP \cite{DBLP:conf/iccv/BhatDGT19}. The top performers in the challenges are improved from the aforementioned trackers, such as C3DT, DeCom, BTT, VITALD in \VIS-SOT2018 \cite{DBLP:conf/eccv/WenZDBLHLCLMNWW18}, ACNT, ATOMFR, ED-ATOM, SMILE and TIOM in \VIS-SOT2019 \cite{DBLP:conf/iccv/DuZWBLHZPWZBS19}, and SMILEv2, LTNMI, ECMMAR, CVP-superdimp, DIMP+SiamRPN, DiMP\_AR and DiMP-101 in \VIS-SOT2020 \cite{DBLP:conf/eccv/FanWDZHLSWDYWZS20}. 
\begin{itemize}
\item \textbf{MDNet} \cite{DBLP:conf/cvpr/NamH16} uses some shared layers and multiple branches of domain-specific layers to construct the network, where the domains correspond to individual training sequences and each branch is responsible for binary classification to identify target in each domain. 
\item \textbf{VITAL} \cite{DBLP:journals/corr/abs-1804-04273} identifies the mask of target that maintains the most robust features over a long temporal span using the adversarial learning.
\item \textbf{CFNet} \cite{DBLP:conf/cvpr/ValmadreBHVT17} is the first attempt to interpret the correlation filter learner as a differentiable layer in a closed-form, which enables learning deep features to be tightly coupled to the correlation filter. 
\item \textbf{TRACA} \cite{DBLP:conf/cvpr/ChoiC0YLJD018} develops a context-aware correlation filter based tracker using multi-expert auto-encoders. The best expert auto-encoder is selected for the target in the tracking phase. 
\item \textbf{ATOM} \cite{DBLP:journals/corr/abs-1811-07628} proposes a tracking architecture formed by dedicated target estimation, which aims to predict the overlap between ground-truth box and estimated box. 
\item \textbf{DiMP} \cite{DBLP:conf/iccv/BhatDGT19} presents a target prediction network, which is derived from a discriminative learning loss by designing a dedicated optimization process for fast convergence. 
\end{itemize}

\subsection{Results and Analysis}
\begin{figure*}[t]
\centering
\includegraphics[width=0.95\linewidth]{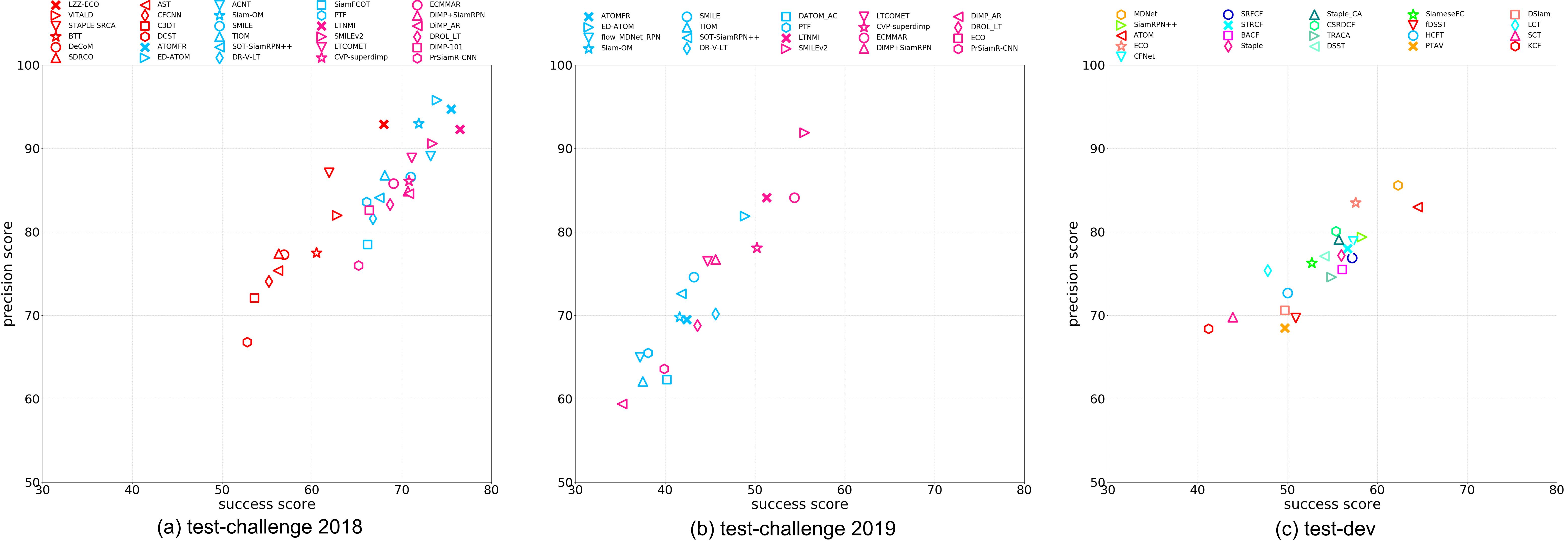}
\caption{The success {\em vs.} precision scores of (a) the top $10$ trackers in \VIS-SOT2018 (red marks), \VIS-SOT2019 (blue marks) and \VIS-SOT2020 (purple marks) on the {\tt test-challenge 2018} set, (b) the top $10$ trackers in \VIS-SOT2019 (blue marks) and \VIS-SOT2020 (purple marks) on the {\tt test-challenge 2019} set, (c) $21$ state-of-the-art trackers on the {\tt test-dev} set.}
\label{fig:sot-res}
\end{figure*}

{\noindent {\bf Results on the {\tt test-challenge} set}.}
We report the success and precision scores of the top $10$ submissions in the \VIS-SOT2018 \cite{DBLP:conf/eccv/WenZDBLHLCLMNWW18}, \VIS-SOT2019 \cite{DBLP:conf/iccv/DuZWBLHZPWZBS19}, and \VIS-SOT2020 \cite{DBLP:conf/eccv/FanWDZHLSWDYWZS20} challenges in Fig. \ref{fig:sot-res}(a) and (b). 

As shown in Fig. \ref{fig:sot-res}(a), we find that most of the correlation filter based methods do not perform well in  \VIS-SOT2018 \cite{DBLP:conf/eccv/WenZDBLHLCLMNWW18}. For example, the winner LZZ-ECO, which combines ECO \cite{DBLP:conf/cvpr/DanelljanBKF17} and YOLOv3 \cite{DBLP:journals/corr/abs-1804-02767}, only produces $68.0$ success score and $92.9$ precision score on the {\tt test-challenge 2018} set. Thus, in the following \VIS-SOT2019 challenge \cite{DBLP:conf/iccv/DuZWBLHZPWZBS19}, researchers shift their focus to the deep neural network based methods. Specifically, ATOMFR integrates squeeze-and-excitation \cite{DBLP:conf/cvpr/HuSS18} into the ATOM \cite{DBLP:journals/corr/abs-1811-07628} tracking framework to capture the interdependencies among feature channels and suppress feature channels that are rarely used in the target size and location prediction, achieving the top accuracy on the {\tt test-challenge 2018} set with $75.5$ success score and $94.7$ precision score. Moreover, in \VIS-SOT2020 \cite{DBLP:conf/eccv/FanWDZHLSWDYWZS20}, researchers combine ATOM \cite{DBLP:journals/corr/abs-1811-07628}, SiamRPN++ \cite{DBLP:journals/corr/abs-1812-11703}, Siam R-CNN \cite{DBLP:conf/cvpr/VoigtlaenderLTL20}, and DiMP \cite{DBLP:conf/iccv/BhatDGT19} to construct the LTNMI tracker, which achieves the best results with $76.5$ success score and $92.3$ precision score on the {\tt test-challenge 2018} set. Notably, multiple trackers ranked in top $10$ are the variants of DiMP \cite{DBLP:conf/iccv/BhatDGT19} in \VIS-SOT2020. 

As shown in Fig. \ref{fig:sot-res}(b), we find that the success and precision scores of the trackers in \VIS-SOT2018 \cite{DBLP:conf/eccv/WenZDBLHLCLMNWW18} are significantly decreased compared with the trackers in \VIS-SOT2019 \cite{DBLP:conf/iccv/DuZWBLHZPWZBS19}, \ie, the best tracker LZZ-ECO in \VIS-SOT2018 produces $68.0$ success score and $92.9$ precision score {\em vs.} the best tracker ED-ATOM in \VIS-SOT2019 produces $48.9$ success score and $81.9$ precision score. This phenomenon demonstrates that the $25$ new collected long-term tracking sequences greatly challenge the performance of the state-of-the-art tracker. Meanwhile, compared with \VIS-SOT2019, we observe significant improvements of tracking accuracy in \VIS-SOT2020. Specifically, the winner SMILEv2 in \VIS-SOT2020 achieves $55.5$ success score and $91.9$ precision score, which surpasses the best tracker ED-ATOM in \VIS-SOT2019 with a large margin. 

Meanwhile, among all submissions in the challenges from 2018 to 2020, some trackers attempt to integrate the state-of-the-art detectors to re-detect the target to deal with drifting problem. For example, in \VIS-SOT2018 \cite{DBLP:conf/eccv/WenZDBLHLCLMNWW18}, LZZ-ECO leverages the detector YOLOv3 \cite{DBLP:journals/corr/abs-1804-02767} to determine the target location if large deformation or camera motion occurs. VITALD trains RefineDet \cite{DBLP:conf/cvpr/ZhangWBLL18} as a reference for the VITAL tracker \cite{DBLP:journals/corr/abs-1804-04273}, \ie, providing reliable target candidates for the target and background classification. In \VIS-SOT2020 \cite{DBLP:conf/eccv/FanWDZHLSWDYWZS20}, DiMP\_AR activates the Faster R-CNN detector \cite{DBLP:journals/pami/RenHG017} to generate target candidates in the video frame if the confidence of tracked target is less than a pre-defined threshold. PrSiamR-CNN is constructed by combining the Siam R-CNN \cite{DBLP:conf/cvpr/VoigtlaenderLTL20} tracker and Faster R-CNN for re-detection.

Another critical engine for performance improvements is the utilization of large-scale datasets, such as MS COCO \cite{DBLP:conf/eccv/LinMBHPRDZ14}, GOT-10k \cite{DBLP:journals/corr/abs-1810-11981}, ImageNet DET/VID \cite{DBLP:journals/ijcv/RussakovskyDSKS15}, LaSOT \cite{DBLP:journals/corr/abs-1809-07845}, TrackingNet \cite{DBLP:conf/eccv/MullerBGAG18}, and YoutubeBB \cite{DBLP:conf/cvpr/RealSMPV17}, in offline training. For example, ED-ATOM achieves the top results in \VIS-SOT2019 \cite{DBLP:conf/iccv/DuZWBLHZPWZBS19}, which is offline trained on ImageNet DET/VID \cite{DBLP:journals/ijcv/RussakovskyDSKS15}, MS COCO \cite{DBLP:conf/eccv/LinMBHPRDZ14}, Got-10k \cite{DBLP:journals/corr/abs-1810-11981}, and LaSOT \cite{DBLP:journals/corr/abs-1809-07845}. Specifically, the ED-ATOM tracker is improved from ATOM \cite{DBLP:journals/corr/abs-1811-07628} by using the low-light image enhancement method \cite{DBLP:conf/iccvw/YingLRWW17} and the online data augmentation scheme \cite{DBLP:conf/eccv/BhatJDKF18}.

In addition, the ensemble strategy is also an effective way to improve the performance. In \VIS-SOT2019 \cite{DBLP:conf/iccv/DuZWBLHZPWZBS19}, Siam-OM adopts ATOM \cite{DBLP:journals/corr/abs-1811-07628} to handle the short-term tracking, and DaSiam \cite{DBLP:conf/eccv/ZhuWLWYH18} uses ResNet to handle the long-term tracking. SIMLE combines two state-of-the-art trackers ATOM \cite{DBLP:journals/corr/abs-1811-07628} and SiamRPN++ \cite{DBLP:journals/corr/abs-1812-11703}, where different features play different roles in tracking process. DR-V-LT integrates the distractor-aware verification network MDNet \cite{DBLP:conf/cvpr/NamH16} into SiamRPN++ \cite{DBLP:journals/corr/abs-1812-11703}, which is robust to the similar objects challenge. In \VIS-SOT2020 \cite{DBLP:conf/eccv/FanWDZHLSWDYWZS20}, the top $3$ performers combine various tracking methods to handle challenges in different scenarios. SMILEv2 is an ensemble of three recent state-of-the-art trackers, \ie, DiMP \cite{DBLP:conf/iccv/BhatDGT19}, SiamMask \cite{DBLP:journals/corr/abs-1812-05050} and SORT \cite{DBLP:conf/icip/WojkeBP17}. Similar to SMILEv2, LTNMI combines four trackers including ATOM \cite{DBLP:journals/corr/abs-1811-07628}, SiamRPN++ \cite{DBLP:journals/corr/abs-1812-11703}, Siam R-CNN \cite{DBLP:conf/cvpr/VoigtlaenderLTL20} and DiMP \cite{DBLP:conf/iccv/BhatDGT19}. ECMMR is constructed by ensemble the results of DiMP \cite{DBLP:conf/iccv/BhatDGT19} and SiamRPN++ \cite{DBLP:journals/corr/abs-1812-11703}. DiMP is used to deal with the target disappearing caused by full occlusion or fast perspective conversion and SiamRPN++ is designed to distinguish distractors in clutter background.

{\noindent {\bf Results on the {\tt test-dev} set}.}
In addition, we evaluate $21$ state-of-the-art trackers on the {\tt test-dev} set in Fig. \ref{fig:sot-res}(c). ATOM \cite{DBLP:journals/corr/abs-1811-07628} (marked as the orange cross in the top-right corner) obtains the best $64.5$ success score and the third best $83.0$ precision score. This is attributed to the network trained offline on large-scale datasets to directly predict the IoU overlap between the target and a bounding box estimate. However, it performs not well in terms of \textit{low resolution} and \textit{out of view}. Besides, MDNet \cite{DBLP:conf/cvpr/NamH16} and SiamRPN++ \cite{DBLP:journals/corr/abs-1812-11703} obtain top $3$ success scores based on large-scale training. In summary, training on large-scale datasets brings significant performance improvement of trackers.

\subsection{Discussion}
As mentioned above, compared with \VIS-SOT2019 \cite{DBLP:conf/iccv/DuZWBLHZPWZBS19}, the tracking accuracy is significantly improved in \VIS-SOT2020 \cite{DBLP:conf/eccv/FanWDZHLSWDYWZS20}. Previously, the correlation filters based trackers \cite{DBLP:conf/cvpr/BertinettoVGMT16,DBLP:conf/cvpr/DanelljanBKF17} are the most popular methods. In recent years, the CNN based trackers \cite{DBLP:journals/corr/abs-1811-07628,DBLP:conf/iccv/BhatDGT19} with online update module quickly dominate this filed. Notably, large-scale training data \cite{DBLP:conf/eccv/LinMBHPRDZ14, DBLP:journals/corr/abs-1810-11981, DBLP:journals/ijcv/RussakovskyDSKS15, DBLP:journals/corr/abs-1809-07845, DBLP:conf/eccv/MullerBGAG18, DBLP:conf/cvpr/RealSMPV17} is crucial for the CNN based trackers. However, long-term tracking is still challenging for the trackers even though the re-detection module \cite{DBLP:journals/corr/abs-1804-02767,DBLP:conf/cvpr/ZhangWBLL18,DBLP:journals/pami/RenHG017} is incorporated. In addition, it still needs extensive efforts to improve the robustness of trackers to the  factors, such as abrupt motion, low resolution, and occlusion, to meet the requirements of real-world applications.

{\noindent \textbf{Abrupt motion}.}
Most of previous SOT methods \cite{DBLP:journals/corr/abs-1812-11703,DBLP:journals/corr/abs-1812-05050} formulate object tracking as the one-shot detection task, which use the bounding box in the first frame as the only exemplar. These methods rely on the pre-set anchor boxes to regress the bounding box of target in consecutive frames. However, the pre-defined anchor boxes can not adapt to various motion patterns and scales of targets, especially when the {\em fast motion} and {\em occlusion} occur. To this end, we can integrate the motion information or re-detection module to improve the accuracy of tracking algorithms.

{\noindent \textbf{Low resolution}} is another challenging factor that greatly affects tracking accuracy. Most of the existing methods \cite{DBLP:journals/corr/abs-1811-07628,DBLP:journals/corr/abs-1812-11703} merely focus on the appearance variations of target region, producing unstable and inaccurate results. We believe that exploiting context information surrounding the target can be helpful to improve the tracking performance, especially for the targets with small scales. 

{\noindent \textbf{Occlusion}} might happen frequently in tracking process, which is the obstacle to produce accurate tracking results. Some previous algorithms \cite{DBLP:journals/tip/DuQLWHL16} attempt to use part-based representations to handle the appearance changes caused by occlusion. Meanwhile, using a re-detection module \cite{DBLP:journals/pami/KalalMM12} is another effective strategy to get rid of occlusion, \ie, the re-detection module is able to detect the target after reappearing in the scenes. In addition, predicting the motion patterns of the target based on its trajectory in history is also a promising direction worth to explore.

\section{MOT Track}
The {\em MOT} track aims to recover the trajectories of objects in video sequences. In the \VIS-VDT2018 challenge \cite{DBLP:conf/eccv/ZhuWDBLHWNCLLMW18}, we divide this track into two sub-tracks depending on whether prior detection results in individual frames are used. In the \VIS-MOT2019 \cite{DBLP:conf/iccv/WenZDBLHZPWZBS19} and \VIS-MOT2020 \cite{DBLP:conf/eccv/FanDWZHLSPSDSXB20} challenges, we merge these two tracks, and do not distinguish submitted algorithms according to whether they use object detection in each video frame as input or not. Notably, this track uses the same data as the {\em VID} track, namely five categories of objects, \ie, {\em pedestrian}, {\em car}, {\em van}, {\em bus}, and {\em truck}, in $96$ video clips.

\subsection{Evaluation Protocol}
For MOT without input detections, we use the metrics in \cite{isvrc-2017} to evaluate the performance. Specifically, we sort the predicted tracklets from algorithms based on the average confidence of the detections of the same identity. If the IoU overlap between the tracklet and the corresponding ground-truth is larger than a threshold, we treat it as a correct one. Following \cite{isvrc-2017}, we average the mean average precision (mAP) per object class over three different thresholds (\ie, $0.25$, $0.50$, and $0.75$) to rank the algorithms. 

For MOT with input detections, we use the CLEAR-MOT metrics in \cite{DBLP:journals/ijcv/DendorferOMSCRR21} for evaluation, \ie, MOTA, MOTP, IDF1, FAF, MT, ML, FP, FN, IDS, and FM. The MOTA metric is the comprehensive metric aggregating three kinds of errors, \ie, FP, FN and IDS. The MOTP metric calculates the average dissimilarity between all true positives and the corresponding ground-truths. The IDF1 metric indicates the ratio of correctly identified detections over the average number of ground truth and computed detections. The FAF metric indicates the average number of false alarms per frame. The FP metric describes the total number of tracker outputs which are the false alarms, and FN is the total number of targets missed by any tracked trajectories in each frame. The IDS metric describes the total number of times that the matched identity of a tracked trajectory changes, while FM is the times that trajectories are disconnected. Both the IDS and FM metrics reflect the accuracy of tracked trajectories. The ML and MT metrics measure the percentage of tracked trajectories less than $20\%$ and more than $80\%$ of the time span based on the ground-truth respectively. More details refer to \cite{DBLP:journals/ijcv/DendorferOMSCRR21}.

\subsection{Review of Multi-Object Tracking Methods}
MOT is a challenging problem due to various factors, such as unreliable detection, long-term occlusion and fast motion, which is a hot topic in recent years. Directly applying single object trackers (\eg, KCF \cite{DBLP:journals/pami/HenriquesC0B15}, CFNet \cite{DBLP:conf/cvpr/ValmadreBHVT17} and DaSiameseRPN \cite{DBLP:conf/eccv/ZhuWLWYH18}) to track multiple objects is a natural way to solve the multi-object tracking task. However, the computation cost increases along with the number of objects in scenes, which is impractical in real-life applications. In this section, we briefly review some representative MOT methods in the challenges and literature. 
 
{\noindent {\bf Tracking-by-detection approach}} is popular in MOT due to its superior performance, which formulates the tracking task as the data association problem. In this scheme, object candidates are detected in each individual frames by using the offline trained detectors. After that, the detected objects are associated to generate object trajectories using various optimization algorithms (\eg, Hungarian algorithm \cite{DBLP:conf/icip/BewleyGORU16,DBLP:conf/icip/WojkeBP17} and max-flow min-cut \cite{DBLP:conf/cvpr/PirsiavashRF11}) based on the appearance or motion informations of objects.
\begin{itemize}
\item {\bf GOG} \cite{DBLP:conf/cvpr/PirsiavashRF11} uses the min-cost flow algorithm to associate the input detections in video frames. The cost function is computed based on the appearance and motion information to determine the number of trajectories and their birth and death states. 
\item {\bf SORT} \cite{DBLP:conf/icip/BewleyGORU16} leverages high-quality detections to use their positions and sizes for both motion estimation and data association. Moreover, Deep SORT \cite{DBLP:conf/icip/WojkeBP17} combines appearance information to track objects through long-term occlusions with less identity switches. 
\item {\bf IOU} \cite{DBLP:conf/avss/BochinskiES17} uses the intersection-over-union (IOU) to measure the similarities of detections in consecutive frames for MOT. 
\item {\bf MOTDT} \cite{DBLP:conf/icmcs/ChenAZS18} handles unreliable detection by collecting candidates from both detection and tracking, and design a scoring function based on CNN to get the optimal selection from a considerable amount of candidates in real-time. 
\item {\bf TrackletNet} \cite{DBLP:journals/corr/abs-1811-07258} constructs a graph model to take the tracklets as the vertices, which exploits the temporal information and greatly reduces the computational complexity. Afterwards, the clustering operation is performed on the graph to generate object trajectories. 
\end{itemize}

{\noindent {\bf Joint tracking and detection approach}} is a recent trend in the MOT field, which integrates object detection and tracking into a unified framework, such as Tracktor \cite{DBLP:conf/iccv/BergmannML19} and FairMOT \cite{DBLP:journals/corr/abs-2004-01888}. In that way, the two tasks can help each other for better performance.
\begin{itemize}
\item {\bf Tracktor} \cite{DBLP:conf/iccv/BergmannML19} uses the bounding box regression of a detector to predict the position of the target in the next frame. In particular, no training or optimization are performed on the tracking data. 
\item {\bf FairMOT} \cite{DBLP:journals/corr/abs-2004-01888} consists of two homogeneous branches to predict pixel-wise objectness scores and re-identification features to obtain high levels of detection and tracking accuracy.
\item {\bf CenterTrack} \cite{DBLP:conf/eccv/ZhouKK20} takes two consecutive frames and a heatmap of prior tracklets as input, and outputs current object centers and tracking offsets to their centers in previous frames.
\end{itemize}

{\noindent {\bf Appearance and motion modeling}.}
In recent years, researchers attempt to leverage deep neural networks to learn discriminative appearance \cite{DBLP:journals/corr/abs-1810-11780,DBLP:conf/iccv/ZhouYCX19,DBLP:conf/cvpr/0004GLL019,DBLP:conf/mm/WangYCLZ18} and motion features \cite{DBLP:conf/icpr/KalalMM10,DBLP:conf/cvpr/IlgMSKDB17,DBLP:conf/cvpr/SunY0K18,DBLP:conf/aaai/MilanRD0S17,DBLP:conf/iccv/SadeghianAS17} of targets. In \VIS-VDT2018 \cite{DBLP:conf/eccv/ZhuWDBLHWNCLLMW18}, \VIS-MOT2019 \cite{DBLP:conf/iccv/WenZDBLHZPWZBS19} and \VIS-MOT2020 \cite{DBLP:conf/eccv/FanDWZHLSPSDSXB20}, some representative methods of appearance modeling are described as follows.
\begin{itemize}
\item {\bf Multi-scale representation}: The deep affinity network \cite{DBLP:journals/corr/abs-1810-11780} associates the detected objects in multiple consecutive frames. Specifically, the multi-scale features are extracted to describe the objects, and the identities of detections at different frames are inferred by analyzing the exhaustive permutations of extracted features.
\item {\bf Onmi-scale representation}: The lightweight OSNet \cite{DBLP:conf/iccv/ZhouYCX19} extracts omni-scale feature representation by the residual block with multiple convolutional streams. Meanwhile, a unified aggregation gate is designed to dynamically fuse multi-scale features.
\item {\bf ReID representation}: A strong ReID model is proposed in \cite{DBLP:conf/cvpr/0004GLL019} without too much extra consumption. Besides, various training tricks are introduced to improve the performance, including warmup learning, random erasing augmentation, label smoothing, smaller last stride, BNNeck and center loss.
\item {\bf Multiple granularity representation}: The multiple granularity network \cite{DBLP:conf/mm/WangYCLZ18} is a multi-branch deep network architecture formed by one branch for global feature representation and two branches for local feature representations, which integrates discriminative information with various granularities. 
\end{itemize}

Besides the appearance information, the motion features are also critical for MOT. Some representative methods are summarized as follows.
\begin{itemize}
\item {\bf Motion patterns} are important low-level cues for MOT, including forward-backward flow \cite{DBLP:conf/icpr/KalalMM10}, KLT\footnote{\url{https://cecas.clemson.edu/\textasciitilde stb/klt/}} and optical flow \cite{DBLP:conf/cvpr/IlgMSKDB17,DBLP:conf/cvpr/SunY0K18}.
\item {\bf Motion networks} are designed to learn the complex long-term temporal dependencies of targets, which are more effective than the predefined motion patterns \cite{DBLP:conf/aaai/MilanRD0S17,DBLP:conf/iccv/SadeghianAS17}. Milan \etal \cite{DBLP:conf/aaai/MilanRD0S17} propose the first end-to-end learning approach for online MOT, without any prior knowledge about target dynamics and clutter distributions. Sadeghian \etal \cite{DBLP:conf/iccv/SadeghianAS17} leverage LSTM networks to track the motion and interactions of targets for longer periods, which is suitable for presence of long-term occlusions.
\end{itemize}

\begin{table}[t]
\centering
\caption{\footnotesize{Comparisons results of the algorithms on the \VIS-MOT dataset using the evaluation protocol in \cite{isvrc-2017}.}}
\setlength{\tabcolsep}{0.5pt}
\scriptsize{
\begin{tabular}{|c|c|ccc|ccccc|}
\hline
Method   &AP &AP$_{@0.25}$ &AP$_{@0.50}$ &AP$_{@0.75}$ &AP$_\text{car}$ &AP$_\text{bus}$ &AP$_\text{trk}$ &AP$_\text{ped}$ &AP$_\text{van}$\\
\hline
		    \multicolumn{10}{c}{\VIS-2018 challenge:}\\			
\hline
Ctrack                  &\textbf{16.12}  &\textbf{22.40} &\textbf{16.26} &\textbf{9.70} &27.74 &\textbf{28.45} &\textbf{8.15} &7.95 &8.31\\
deep-sort\_d2           &10.47  &17.26 &9.40 &4.75 &\textbf{29.14} &2.38 &3.46 &7.12 &\textbf{10.25}\\
MAD                     &7.27  &12.72  &7.03 &2.07  &16.23 &1.65 &2.85 &\textbf{14.16} &1.46\\
\hline
		    \multicolumn{10}{c}{\VIS-2019 challenge:}\\			
\hline
    DBAI-Tracker & \textbf{43.94} & \textbf{57.32} & \textbf{45.18} & 29.32 & \textbf{55.13} & \textbf{44.97} & \textbf{42.73} & 31.01 & \textbf{45.85} \\
    TrackKITSY & 39.19 & 48.83 & 39.36 & \textbf{29.37} & 54.92 & 29.05 & 34.19 & \textbf{36.57} & 41.20 \\
    Flow-Tracker & 30.87 & 41.84 & 31.00    & 19.77 & 48.44 & 26.19 & 29.50  & 18.65 & 31.56 \\
    HMTT  & 28.67 & 39.05 & 27.88 & 19.08 & 44.35 & 30.56 & 18.75 & 26.49 & 23.19 \\
    TNT\_DRONE & 27.32 & 35.09 & 26.92 & 19.94 & 38.06 & 22.65 & 33.79 & 12.62 & 29.46 \\
    GGDTRACK & 23.09 & 31.01 & 22.70  & 15.55 & 35.45 & 28.57 & 11.90  & 17.20  & 22.34 \\
    IITD\_DeepSort & 13.88 & 23.19 & 12.81 & 5.64  & 32.20  & 8.83  & 6.61  & 18.61 & 3.16 \\
    T\&D-OF & 12.37 & 17.74 & 12.94 & 6.43  & 23.31 & 22.02 & 2.48  & 9.59  & 4.44 \\
    SCTrack & 10.09 & 14.95 & 9.41  & 5.92  & 18.98 & 17.86 & 4.86  & 5.20   & 3.58 \\
    VCLDAN & 7.50   & 10.75 & 7.41  & 4.33  & 21.63 & 0.00     & 4.92  & 10.94 & 0.00 \\
\hline
		    \multicolumn{10}{c}{\VIS-2020 challenge:}\\			
\hline
		COFE & \textbf{61.88} & 64.99 &\textbf{62.00} & \textbf{58.65} & \textbf{79.09} & \textbf{65.26} & \textbf{50.91} & \textbf{56.87} & \textbf{57.26} \\
		SOMOT  & 57.65 & \textbf{70.06} & 60.13 & 42.75 & 68.52 & 62.10 & 47.98 & 54.94 & 54.69 \\
		PAS & 50.80 & 62.24 & 50.74 & 39.43 & 62.59 & 50.59 & 42.18 & 44.34 & 54.30 \\
		Deepsort  & 42.11 & 58.82 & 42.64 & 24.86 & 55.06 & 43.18 & 41.30 & 29.10 & 41.88 \\
		YOLO-TRAC & 42.10 & 52.94 & 41.86 & 31.49 & 52.81 & 48.98 & 39.17 & 28.92 & 40.59 \\
		VDCT  & 35.76 & 45.86 & 35.46 & 25.96 & 56.94 & 24.62 & 28.16 & 34.00 & 35.06 \\
		CRCNN+IOU & 27.23 & 36.14 & 28.25 & 17.31 & 49.56 & 16.27 & 30.18 & 10.78 & 29.36 \\
		HTC+IOU  & 26.46 & 34.39 & 27.43 & 17.57 & 51.18 & 19.05 & 21.55 & 10.77 & 29.76 \\
		HR-GNN & 19.54 & 26.52 & 19.67 & 12.42 & 37.72 & 15.48 & 9.98  & 18.87 & 15.65 \\
		TNT  & 6.55  & 10.93 & 7.00  & 1.70  & 1.88  & 19.51 & 2.07  & 1.96  & 7.32 \\
\hline		    		
		    \multicolumn{10}{c}{\VIS-dev:}\\	
\hline		    		
    GOG \cite{DBLP:conf/cvpr/PirsiavashRF11} & \textbf{5.14} & \textbf{11.02} & 3.25 & 1.14 & \textbf{13.70} & \textbf{3.09} & 1.94 & 3.08 & 3.87 \\
    IOUT \cite{DBLP:conf/avss/BochinskiES17} & 4.34 & 8.32 & \textbf{3.29} & 1.40 & 10.90 & 2.15 & \textbf{2.53} & 1.98 & \textbf{4.11}  \\
    SORT \cite{DBLP:conf/icip/WojkeBP17} & 3.37 & 5.78 & 2.82 & \textbf{1.50} & 8.30 & 1.04 & 2.47 & 0.95 & 4.06  \\
    MOTDT \cite{DBLP:conf/icmcs/ChenAZS18} & 1.22 & 2.43 & 0.92 & 0.30 & 0.36 & 0.00 & 0.15 & \textbf{5.08} & 0.49  \\
\hline
\end{tabular}}
\label{tab:mot-res:a}
\end{table}

\begin{table}[t]
\centering
\caption{\footnotesize{Comparisons results of the algorithms on the \VIS-MOT dataset using the CLEAR-MOT evaluation protocol \cite{DBLP:journals/ijcv/DendorferOMSCRR21}.}}
\setlength{\tabcolsep}{2pt}
\scriptsize{
\begin{tabular}{|c|cccc|cccccc|}
\hline
Method   &MOTA &MOTP &IDF1 &FAF &MT &ML &FP &FN &IDS &FM\\
\hline
		    \multicolumn{11}{c}{\VIS-2018 challenge:}\\	
\hline
TrackCG  &\textbf{42.6} &74.1 &\textbf{58.0} &0.86 &323 &395 &14722 &68060 &779  &3717 \\
V-IOU    &40.2 &74.9 &56.1 &0.76 &297 &514 &11838 &74027 &\textbf{265} &1380 \\
GOG\_EOC &36.9 &\textbf{75.8} &46.5 &\textbf{0.29} &205 &589 &\textbf{5445} &86399 &354  &\textbf{1090} \\
SCTrack  &35.8  &75.6 &45.1 &0.39 &211 &550 &7298 &85623 &798 &2042 \\
FRMOT    &33.1 &73.0 &50.8 &1.15 &254 &463 &21736 &74953 &1043 &2534 \\
Ctrack   &30.8 &73.5 &51.9 &1.95 &\textbf{369} &\textbf{375} &36930 &\textbf{62819} &1376 &2190 \\
\hline
		    \multicolumn{11}{c}{\VIS-dev:}\\	
\hline
GOG \cite{DBLP:conf/cvpr/PirsiavashRF11} & \textbf{28.7} & \textbf{76.1} & 36.4 & \textbf{0.78} & 346 & 836 & \textbf{17706} & 144657 & \textbf{1387} & \textbf{2237} \\
IOUT \cite{DBLP:conf/avss/BochinskiES17} & 28.1 & 74.7 & \textbf{38.9} & 1.60 & 467 & 670 & 36158 & 126549 & 2393 & 3829 \\
SORT \cite{DBLP:conf/icip/WojkeBP17} & 14.0 & 73.2 & 38.0 & 3.57 & \textbf{506} & \textbf{545} & 80845 & \textbf{112954} & 3629 & 4838 \\
MOTDT \cite{DBLP:conf/icmcs/ChenAZS18} & -0.8 & 68.5 & 21.6 & 1.97 & 87 & 1196 & 44548 & 185453 & 1437 & 3609 \\
\hline
\end{tabular}}
\label{tab:mot-res:b}
\end{table}

\subsection{Results and Analysis}
{\noindent {\bf Results on the {\tt test-challenge} set}.}
We report the evaluation results of top $10$ MOT methods in \VIS-VDT2018 \cite{DBLP:conf/eccv/ZhuWDBLHWNCLLMW18}, \VIS-MOT2019 \cite{DBLP:conf/iccv/WenZDBLHZPWZBS19} and \VIS-MOT2020 \cite{DBLP:conf/eccv/FanDWZHLSPSDSXB20} based on the evaluation protocols \cite{isvrc-2017} and \cite{DBLP:journals/ijcv/DendorferOMSCRR21} in Table \ref{tab:mot-res:a} and \ref{tab:mot-res:b}, respectively. In summary, the MOT performance highly relies on three aspects, \ie, input object detection quality, motion models and appearance models.

As shown in Table \ref{tab:mot-res:a}, the submissions in \VIS-MOT2019 \cite{DBLP:conf/iccv/WenZDBLHZPWZBS19} achieve significant improvement than that in \VIS-VDT2018 \cite{DBLP:conf/eccv/ZhuWDBLHWNCLLMW18}. For example, the best DBAI-Tracker in \VIS-MOT2019 \cite{DBLP:conf/iccv/WenZDBLHZPWZBS19} achieves $25\%$ higher AP score than the best tracker Ctrack in \VIS-VDT2018 \cite{DBLP:conf/eccv/ZhuWDBLHWNCLLMW18}. This is partially attributed to the powerful Cascade R-CNN \cite{DBLP:conf/cvpr/CaiV18} detector, which provides more accurate input detections than Faster R-CNN \cite{DBLP:journals/pami/RenHG017} and RetinaNet \cite{DBLP:conf/iccv/LinGGHD17}.

Meanwhile, exploiting motion information is an effective way to improve the performance. In \VIS-VDT2018 \cite{DBLP:conf/eccv/ZhuWDBLHWNCLLMW18}, Ctrack achieves the best AP score $16.12\%$ by aggregating the prediction events in grouped targets and stitching the tracks by temporal constraints. In this way, Ctrack can recover the trajectories of occluded targets. Although deep-sort\_d2 and MAD use more powerful object detectors RetinaNet \cite{DBLP:conf/iccv/LinGGHD17} and YOLOv3 \cite{DBLP:journals/corr/abs-1804-02767}, they fail to make full use of the motion patterns within grouped targets, resulting in inferior results. In \VIS-MOT2019 \cite{DBLP:conf/iccv/WenZDBLHZPWZBS19}, three out of the top four trackers (DBAI-Tracker, Flow-Tracker, and HMTT) integrate additional temporal modules such as optical flow \cite{DBLP:conf/cvpr/IlgMSKDB17,DBLP:conf/cvpr/SunY0K18} to complete association, demonstrating the critical role of motion information in the tracking task. 

In \VIS-MOT2020 \cite{DBLP:conf/eccv/FanDWZHLSPSDSXB20}, the AP scores of the submitted algorithms are further improved compared with \VIS-MOT2019 \cite{DBLP:conf/iccv/WenZDBLHZPWZBS19}. The top three methods are COFE, SOMOT and PAS. Besides the strong Cascade R-CNN \cite{DBLP:conf/cvpr/CaiV18} detector and motion modeling, the re-identification model \cite{DBLP:conf/iccv/ZhouYCX19,DBLP:conf/mm/WangYCLZ18,DBLP:conf/cvpr/0004GLL019} contributes a great deal to the tracking performance. For example, SOMOT achieves better performance by considering both global and local appearance \cite{DBLP:conf/mm/WangYCLZ18}, compared to PAS that only relies on global features \cite{DBLP:conf/cvpr/0004GLL019}. Without the re-identification models, Deepsort and YOLO-TRAC produce inferior AP score of $42.1\%$. It is worth mentioning that the best tracker COFE employs a coarse-category training strategy. Specifically, it first performs multi-object tracking for each coarse category, \eg, fine-grained classes such as ``van'', ``bus'' and ``car'' are grouped into a coarse category ``vehicle''. After that, the fine-grained class labels of targets are voted by the class labels of detections on the trajectories. For example, if the length of target trajectory is $20$, including $10$ detections classified as ``car'', $6$ detections classified as ``bus'', and $4$ detections classified as ``van'', the target is voted as the fine-grained class ``car''.

For the sub-track in \VIS-VDT2018 \cite{DBLP:conf/eccv/ZhuWDBLHWNCLLMW18} with input detections generated by Faster R-CNN \cite{DBLP:journals/pami/RenHG017}, TrackCG achieves the best MOTA and IDF1 scores. V-IOU produces slightly inferior MOTA and IDF1 scores than TrackCG, but lower IDS score. It associates object detections in consecutive frames based on the spatial intersection-over-union overlap. We speculate that the overlapping measurement is reliable enough for tracking in drone captured videos, which do not contain large displacements of objects in consecutive frames. GOG\_EOC achieves the best FAF, FP and FM scores, which uses the overlap and context harmony degree to measure the similarities of detections. SCTrack designs the color correlation costs to maintain object identities. However, the color information is not reliable enough, resulting in inferior MOTA score. FRMOT is an online tracker using the Hungarian algorithm to associate detections, producing relative large IDS and FM scores.

{\noindent {\bf Results on the {\tt test-dev} set}.}
Besides, we evaluate $4$ multi-object tracking on the {\tt test-dev} set with the evaluation protocols \cite{isvrc-2017} and \cite{DBLP:journals/ijcv/DendorferOMSCRR21}, shown in Table \ref{tab:mot-res:a} and \ref{tab:mot-res:b}, respectively. Notably, FPN \cite{DBLP:conf/cvpr/LinDGHHB17} is used to generate object detections in individual frames for the sub-track using prior input detections. GOG \cite{DBLP:conf/cvpr/PirsiavashRF11} and IOUT \cite{DBLP:conf/avss/BochinskiES17} benefit from global information of whole sequences and spatial overlap between frame detections, achieving the best tracking results in terms of both evaluation protocols \cite{isvrc-2017} and \cite{DBLP:journals/ijcv/DendorferOMSCRR21}. Deep-SORT \cite{DBLP:conf/icip/WojkeBP17} approximates the inter-frame displacements of each object with a linear constant velocity model, which is independent of object categories and camera motion, significantly degrading its performance. MOTDT \cite{DBLP:conf/icmcs/ChenAZS18} computes the similarities between objects using appearance model trained on other large-scale person re-identification datasets without fine-tuning, leading to inferior accuracy.

\subsection{Discussion}
Most of the MOT methods formulate the tracking task as the data association problem, which aims to associate object detections in sequential frames to generate object trajectories. Thus, the accuracy of object detection significantly influence the performance of MOT. Some submitted MOT methods focus on obtaining strong detections \cite{DBLP:journals/pami/RenHG017,DBLP:conf/iccv/LinGGHD17,DBLP:conf/cvpr/CaiV18} and fine-grained feature representations \cite{DBLP:conf/iccv/ZhouYCX19,DBLP:conf/mm/WangYCLZ18,DBLP:conf/cvpr/0004GLL019}, which is the key to achieve the state-of-the-art results. However, the two-stage strategy, \ie, detection and association, is suboptimal due to the separation of these two tasks. Intuitively, integrating object detection and tracking into a unified framework is intuitive to improve the performance. First, the end-to-end model has lower computational complexity and higher efficiency. Second, these two tasks can share information to improve the accuracy. In the following, we discuss two potential research directions to further boost the MOT performance.

{\flushleft \textbf{Similarity estimation}.} For the data association problem, similarity computation between different detections in individual frames is crucial for the tracking performance. The appearance and motion information should be considered in computing the similarities. For example, a Siamese network offline trained on the ImageNet VID dataset  \cite{DBLP:journals/ijcv/RussakovskyDSKS15} can be used to exploit temporal discriminative features of objects. It can be fine-tuned in tracking process to further improve the accuracy. Meanwhile, several low-level and mid-level motion features are also effective and useful for the MOT algorithms, such as KLT and optical flow.

{\flushleft \textbf{Scene understanding}} is another effective way to improve the MOT performance. For example, based on the scene understanding module, we can infer the enter or exit ports in the scenes as a strong priori for the trackers to distinguish occlusion, termination, or re-appearing of the targets. Meanwhile, the tracker can also suppress false trajectories based on general knowledge, \eg, the vehicles are only driven on the road rather on the building. In summary, this area is worth further studying.

\section{Conclusion and Future Research}
We introduce a new large-scale benchmark, {\bf \VIS}, to facilitate the research of object detection and tracking on drone captured imagery. With over $6,000$ worker hours, a vast collection of object instances are gathered, annotated, and organized to drive the advancement of object detection and tracking algorithms. Meanwhile, several representative submitted detection and tracking methods are reviewed, where the best submissions in the four tracks are still short for real applications. From both research and application perspectives, this work discusses some under-developed but critical future issues of object detection and tracking on drone captured imagery. It would shed some light into potential future research directions on drone-based object detection and tracking.

{\flushleft {\bf Relations between object detection and tracking.}}
Early on, object detection is generally used as the first step to generate candidate object proposals, and object tracking algorithms are required to associate the object proposals to recover the trajectories, \ie, the so-called ``tracking-by-detection'' framework. Specifically, early object detection methods rely on the sliding-window paradigm and design hand-craft features and classifiers on dense image grids to detect objects, such as Adaboost \cite{DBLP:conf/cvpr/ViolaJ01}, DPM \cite{DBLP:conf/cvpr/FelzenszwalbMR08}, and ACF \cite{DBLP:journals/pami/DollarABP14}. Deep convolutional network dominates the detection field in recent years due to its powerful representation capability, including Faster R-CNN \cite{DBLP:journals/pami/RenHG017}, Cascade R-CNN \cite{DBLP:conf/cvpr/CaiV18}, YOLOv3 \cite{DBLP:journals/corr/abs-1804-02767}, RefineDet \cite{DBLP:conf/cvpr/ZhangWBLL18} and CenterNet \cite{DBLP:journals/corr/abs-1904-07850}. Meanwhile, for object tracking, most of early methods use the tracking-by-detection paradigm, such as Hungarian algorithm \cite{DBLP:conf/eccv/HuangWN08}, min-cost flow \cite{DBLP:conf/bmvc/WangF15}, and hypergraph optimization \cite{DBLP:conf/aaai/WenDLBL19}, which takes object detection results in individual frames as input, and design algorithms to associate object detections in each frame by exploiting temporal information to complete object tracking.  

Intuitively, tracking systems clearly benefit from accurate object detections, while are effective to help detectors to produce accurate results by exploiting the temporal information. Some researchers attempt to design the end-to-end frameworks \cite{DBLP:conf/iccv/FeichtenhoferPZ17,DBLP:conf/cvpr/VoigtlaenderKOL19,DBLP:conf/cvpr/LuRVH20,DBLP:conf/cvpr/PangLZLL20} based on deep neural networks to solve object detection and tracking jointly. In this way, object detection and tracking can help each other to boost the performance. We believe that the joint detection and tracking are especially important for the drone-captured videos. Since the scales of objects are extremely tiny in drone-captured videos, heavily relying on appearance information in traditional methods is less effective. Combining detection and tracking in a unified framework to take advantage of the spatio-temporal context information is worth to pursue. 

{\flushleft \textbf{Performance evaluation.}}
In various scenarios, different factors have different importance for detection and tracking. For example, in surveillance scenarios, the accuracies of detections and trajectories are of great importance. Thus we should put more weights on the metrics based on id switches, trajectory fragmentations, and false positives. In autonomous driving scenarios, the false negatives are of more concern than other factors. Building an appropriate evaluation protocol to meet the requirements in different scenarios is an urgent task in object detection and tracking. 

{\flushleft \textbf{Effectiveness and efficiency.}}
Effectiveness and efficiency are both important aspects for algorithms in real applications. For example, the computation resources are limited on drone platform, making it impossible to deploy a large model on such edge devices. In the \VIS-Challenges from 2018 to 2020, all submissions focus more on accuracy than efficiency. To promote the developments of algorithms for real-world applications, we plan to consider the running efficiency in the evaluation process in the following \VIS-Challenges. Recently, the automated neural architecture search (AutoNAS) is a hot topic in research field. It is potentially effective to introduce the AutoNAS technologies \cite{DBLP:journals/corr/abs-1911-13053,DBLP:conf/cvpr/YanDZTWF21} to search the model architectures by considering both the effectiveness and efficiency for the drone-based applications. Specifically, for different drone platforms, we need to consider the specific computational power and memory request in the architecture searching process, while maintaining the accuracy. This direction is relatively less developed, and we need to put more efforts on it.

{\flushleft \textbf{Unsolved problems.}}
There still exists several challenges in drone-captured video sequences, such as viewpoint change, motion blur, abrupt motion and small objects. Although the submitted object detection and tracking methods produce promising results in the deep learning era, they are still prone to fail in the aforementioned challenges. This is because most of previous object detection and tracking methods are on the rack to exploit the appearance information to improve the performance. Actually, it is extremely difficult to deal with the challenging cases solely relying on appearance information, \eg, detecting blurred/occluded objects in videos, or tracking multiple objects with similar appearance.

To that end, a few attempts exploit the motion information to improve the accuracy, such as high-order motion constraints \cite{DBLP:conf/cvpr/Collins12}, tensor power iterations \cite{DBLP:conf/cvpr/ShiLHYX14}, and dense structure search on hypergraphs \cite{DBLP:conf/aaai/WenDLBL19}. Although these methods are effective in some scenarios, it is still much room for improvements to meet the requirements of the applications on drones. Constructing a reliable motion model to fully exploit the motion information is a promising direction for future research.

On the other hand, current submissions handle with small objects by using multi-scale representations \cite{DBLP:conf/cvpr/LinDGHHB17} and data augmentation strategies \cite{DBLP:conf/nips/SinghND18}. However, it is difficult to learn effective representations of small objects based on the generic ImageNet or MS-COCO datasets, resulting in inferior performance on our dataset. To improve the accuracy of detectors, it is urgent to collect large-scale datasets and benchmarks for small objects. Moreover, anchor-free object detection methods \cite{DBLP:journals/corr/abs-1904-07850,DBLP:journals/corr/abs-1904-01355} focus on detecting objects by points, which may be another research direction for small object detection.

\section*{Acknowledgements}
We would like to thank Jiayu Zheng and Tao Peng for valuable and constructive suggestions to improve the quality of this paper. 

\small
{
\bibliographystyle{IEEEtran}
\bibliography{reference-short}

\begin{thebibliography}{100}
\providecommand{\url}[1]{#1}
\csname url@samestyle\endcsname
\providecommand{\newblock}{\relax}
\providecommand{\bibinfo}[2]{#2}
\providecommand{\BIBentrySTDinterwordspacing}{\spaceskip=0pt\relax}
\providecommand{\BIBentryALTinterwordstretchfactor}{4}
\providecommand{\BIBentryALTinterwordspacing}{\spaceskip=\fontdimen2\font plus
\BIBentryALTinterwordstretchfactor\fontdimen3\font minus
  \fontdimen4\font\relax}
\providecommand{\BIBforeignlanguage}[2]{{%
\expandafter\ifx\csname l@#1\endcsname\relax
\typeout{** WARNING: IEEEtran.bst: No hyphenation pattern has been}%
\typeout{** loaded for the language `#1'. Using the pattern for}%
\typeout{** the default language instead.}%
\else
\language=\csname l@#1\endcsname
\fi
#2}}
\providecommand{\BIBdecl}{\relax}
\BIBdecl

\bibitem{drone-report}
G.~V. Research, ``Commercial drone market size, share \& trends analysis report
  by product (fixed-wing, rotary blade, hybrid), by application, by end-use, by
  region, and segment forecasts, 2021 - 2028,''
  \url{https://www.reportlinker.com/p06068219/?utm_source=GNW}, 2021.

\bibitem{DBLP:conf/eccv/MuellerSG16}
M.~Mueller, N.~Smith, and B.~Ghanem, ``A benchmark and simulator for {UAV}
  tracking,'' in \emph{ECCV}, 2016, pp. 445--461.

\bibitem{DBLP:conf/fgr/KalraSN0VS19}
I.~Kalra, M.~Singh, S.~Nagpal, R.~Singh, M.~Vatsa, and P.~B. Sujit,
  ``Dronesurf: Benchmark dataset for drone-based face recognition,'' in
  \emph{FG}, 2019, pp. 1--7.

\bibitem{DBLP:conf/eccv/DuQYYDLZHT18}
D.~Du, Y.~Qi, H.~Yu, Y.~Yang, K.~Duan, G.~Li, W.~Zhang, Q.~Huang, and Q.~Tian,
  ``The unmanned aerial vehicle benchmark: Object detection and tracking,'' in
  \emph{ECCV}, 2018, pp. 375--391.

\bibitem{DBLP:conf/iccv/HsiehLH17}
M.~Hsieh, Y.~Lin, and W.~H. Hsu, ``Drone-based object counting by spatially
  regularized regional proposal network,'' in \emph{ICCV}, 2017.

\bibitem{DBLP:conf/eccv/RobicquetSAS16}
A.~Robicquet, A.~Sadeghian, A.~Alahi, and S.~Savarese, ``Learning social
  etiquette: Human trajectory understanding in crowded scenes,'' in
  \emph{ECCV}, 2016, pp. 549--565.

\bibitem{DBLP:conf/eccv/ZhuWDBLHNCLLMWW18}
P.~Zhu, L.~Wen, D.~Du, X.~Bian, H.~Ling, Q.~Hu, and \textit{et al.},
  ``Visdrone-det2018: The vision meets drone object detection in image
  challenge results,'' in \emph{{ECCV} Workshops}, 2018, pp. 437--468.

\bibitem{DBLP:conf/eccv/WenZDBLHLCLMNWW18}
L.~Wen, P.~Zhu, D.~Du, X.~Bian, H.~Ling, Q.~Hu, and \textit{et al.},
  ``Visdrone-sot2018: The vision meets drone single-object tracking challenge
  results,'' in \emph{{ECCV} Workshops}, 2018, pp. 469--495.

\bibitem{DBLP:conf/eccv/ZhuWDBLHWNCLLMW18}
P.~Zhu, L.~Wen, D.~Du, X.~Bian, H.~Ling, Q.~Hu, and \textit{et al.},
  ``Visdrone-vdt2018: The vision meets drone video detection and tracking
  challenge results,'' in \emph{{ECCV} Workshops}, 2018, pp. 496--518.

\bibitem{DBLP:conf/iccv/DuZWBLHPDET19}
D.~Du, P.~Zhu, L.~Wen, X.~Bian, H.~Ling, Q.~Hu, and \textit{et al.},
  ``Visdrone-det2019: The vision meets drone object detection in image
  challenge results,'' in \emph{{ICCV} Workshops}, 2019.

\bibitem{DBLP:conf/iccv/DuZWBLHZPWZBS19}
D.~Du, P.~Zhu, L.~Wen, X.~Bian, H.~Ling, Q.~Hu, and \textit{et al.},
  ``Visdrone-sot2019: The vision meets drone single object tracking challenge
  results,'' in \emph{{ICCV} Workshops}, 2019.

\bibitem{DBLP:conf/iccv/ZhuDWBLHPZWZBS19}
P.~Zhu, D.~Du, L.~Wen, X.~Bian, H.~Ling, Q.~Hu, and \textit{et al.},
  ``Visdrone-vid2019: The vision meets drone object detection in video
  challenge results,'' in \emph{{ICCV} Workshops}, 2019.

\bibitem{DBLP:conf/iccv/WenZDBLHZPWZBS19}
L.~Wen, P.~Zhu, D.~Du, X.~Bian, H.~Ling, Q.~Hu, and \textit{et al.},
  ``Visdrone-mot2019: The vision meets drone multiple object tracking challenge
  results,'' in \emph{{ICCV} Workshops}, 2019.

\bibitem{DBLP:conf/eccv/DuWZFHLSPASPJDL20}
D.~Du, L.~Wen, P.~Zhu, H.~Fan, Q.~Hu, H.~Ling, M.~Shah, J.~Pan, and \textit{et
  al.}, ``Visdrone-det2020: The vision meets drone object detection in image
  challenge results,'' in \emph{ECCV Workshops}, vol. 12538, 2020, pp.
  692--712.

\bibitem{DBLP:conf/eccv/FanDWZHLSPSDSXB20}
H.~Fan, D.~Du, L.~Wen, P.~Zhu, Q.~Hu, H.~Ling, M.~Shah, J.~Pan, and \textit{et
  al.}, ``Visdrone-mot2020: The vision meets drone multiple object tracking
  challenge results,'' in \emph{ECCV Workshops}, vol. 12538, 2020, pp.
  713--727.

\bibitem{DBLP:conf/eccv/FanWDZHLSWDYWZS20}
H.~Fan, L.~Wen, D.~Du, P.~Zhu, Q.~Hu, H.~Ling, M.~Shah, and \textit{et al.},
  ``Visdrone-sot2020: The vision meets drone single object tracking challenge
  results,'' in \emph{ECCV Workshops}, vol. 12538, 2020, pp. 728--749.

\bibitem{pascal-voc-2012}
M.~Everingham, L.~Van~Gool, C.~K.~I. Williams, J.~Winn, and A.~Zisserman, ``The
  {PASCAL} {V}isual {O}bject {C}lasses {C}hallenge 2012 {(VOC2012)}
  {R}esults,''
  http://www.pascal-network.org/challenges/VOC/voc2012/workshop/index.html,
  2012.

\bibitem{DBLP:journals/ijcv/RussakovskyDSKS15}
O.~Russakovsky, J.~Deng, H.~Su, J.~Krause, S.~Satheesh, S.~Ma, Z.~Huang,
  A.~Karpathy, A.~Khosla, M.~S. Bernstein, A.~C. Berg, and F.~Li, ``Imagenet
  large scale visual recognition challenge,'' \emph{IJCV}, vol. 115, no.~3, pp.
  211--252, 2015.

\bibitem{DBLP:conf/eccv/LinMBHPRDZ14}
T.~Lin, M.~Maire, S.~J. Belongie, J.~Hays, P.~Perona, D.~Ramanan,
  P.~Doll{\'{a}}r, and C.~L. Zitnick, ``Microsoft {COCO:} common objects in
  context,'' in \emph{ECCV}, 2014, pp. 740--755.

\bibitem{DBLP:conf/eccv/MundhenkKSB16}
T.~N. Mundhenk, G.~Konjevod, W.~A. Sakla, and K.~Boakye, ``A large contextual
  dataset for classification, detection and counting of cars with deep
  learning,'' in \emph{ECCV}, 2016, pp. 785--800.

\bibitem{DBLP:conf/cvpr/XiaBDZBLDPZ18}
G.~Xia, X.~Bai, J.~Ding, Z.~Zhu, S.~J. Belongie, J.~Luo, M.~Datcu, M.~Pelillo,
  and L.~Zhang, ``{DOTA:} {A} large-scale dataset for object detection in
  aerial images,'' in \emph{CVPR}, 2018, pp. 3974--3983.

\bibitem{DBLP:conf/mm/MandalKV20}
M.~Mandal, L.~K. Kumar, and S.~K. Vipparthi, ``{MOR-UAV:} {A} benchmark dataset
  and baselines for moving object recognition in {UAV} videos,'' in \emph{{ACM}
  International Conference on Multimedia}, 2020, pp. 2626--2635.

\bibitem{DBLP:journals/corr/WenDCLCQLYL15}
L.~Wen, D.~Du, Z.~Cai, Z.~Lei, M.~Chang, H.~Qi, J.~Lim, M.~Yang, and S.~Lyu,
  ``{UA-DETRAC:} {A} new benchmark and protocol for multi-object detection and
  tracking,'' \emph{Computer Vision and Image Understanding}, vol. 193, p.
  102907, 2020.

\bibitem{DBLP:journals/ijcv/DendorferOMSCRR21}
P.~Dendorfer, A.~Osep, A.~Milan, K.~Schindler, D.~Cremers, I.~Reid, S.~Roth,
  and L.~Leal{-}Taix{\'{e}}, ``Motchallenge: {A} benchmark for single-camera
  multiple target tracking,'' \emph{IJCV}, vol. 129, no.~4, pp. 845--881, 2021.

\bibitem{DBLP:conf/cvpr/BarekatainMSMNM17}
M.~Barekatain, M.~Mart{\'{\i}}, H.~Shih, S.~Murray, K.~Nakayama, Y.~Matsuo, and
  H.~Prendinger, ``Okutama-action: An aerial view video dataset for concurrent
  human action detection,'' in \emph{CVPRWorkshops}, 2017, pp. 2153--2160.

\bibitem{DBLP:journals/pami/WuLY15}
Y.~Wu, J.~Lim, and M.~Yang, ``Object tracking benchmark,'' \emph{TPAMI},
  vol.~37, no.~9, pp. 1834--1848, 2015.

\bibitem{DBLP:conf/eccv/KristanLMFPCVHL16}
M.~Kristan, A.~Leonardis, J.~Matas, M.~Felsberg, R.~P. Pflugfelder, L.~Cehovin,
  T.~Voj{\'{\i}}r, G.~H{\"{a}}ger, A.~Lukezic, G.~Fern{\'{a}}ndez, and
  \textit{et al.}, ``The visual object tracking {VOT2016} challenge results,''
  in \emph{ECCVWorkshops}, 2016, pp. 777--823.

\bibitem{DBLP:conf/aaai/LiY17}
S.~Li and D.~Yeung, ``Visual object tracking for unmanned aerial vehicles: {A}
  benchmark and new motion models,'' in \emph{AAAI}, 2017, pp. 4140--4146.

\bibitem{DBLP:journals/corr/abs-1810-11981}
L.~Huang, X.~Zhao, and K.~Huang, ``Got-10k: {A} large high-diversity benchmark
  for generic object tracking in the wild,'' \emph{{IEEE} Trans. Pattern Anal.
  Mach. Intell.}, vol.~43, no.~5, pp. 1562--1577, 2021.

\bibitem{DBLP:conf/eccv/MullerBGAG18}
M.~M{\"{u}}ller, A.~Bibi, S.~Giancola, S.~Al{-}Subaihi, and B.~Ghanem,
  ``Trackingnet: {A} large-scale dataset and benchmark for object tracking in
  the wild,'' in \emph{ECCV}, 2018, pp. 310--327.

\bibitem{DBLP:journals/corr/abs-2003-06994}
P.~Zhu, J.~Zheng, D.~Du, L.~Wen, Y.~Sun, and Q.~Hu, ``Multi-drone based single
  object tracking with agent sharing network,'' \emph{TCSVT}, 2020.

\bibitem{DBLP:journals/corr/abs-1809-07845}
H.~Fan, L.~Lin, F.~Yang, P.~Chu, G.~Deng, S.~Yu, H.~Bai, Y.~Xu, C.~Liao, and
  H.~Ling, ``Lasot: {A} high-quality benchmark for large-scale single object
  tracking,'' in \emph{CVPR}, 2019, pp. 5374--5383.

\bibitem{DBLP:journals/corr/abs-2101-08466}
N.~Jiang, K.~Wang, X.~Peng, X.~Yu, Q.~Wang, J.~Xing, G.~Li, J.~Zhao, G.~Guo,
  and Z.~Han, ``Anti-uav: {A} large multi-modal benchmark for {UAV} tracking,''
  \emph{CoRR}, vol. abs/2101.08466, 2021.

\bibitem{DBLP:conf/cvpr/GeigerLU12}
A.~Geiger, P.~Lenz, and R.~Urtasun, ``Are we ready for autonomous driving? the
  {KITTI} vision benchmark suite,'' in \emph{CVPR}, 2012, pp. 3354--3361.

\bibitem{DBLP:conf/eccv/RistaniSZCT16}
E.~Ristani, F.~Solera, R.~S. Zou, R.~Cucchiara, and C.~Tomasi, ``Performance
  measures and a data set for multi-target, multi-camera tracking,'' in
  \emph{ECCVWorkshops}, 2016, pp. 17--35.

\bibitem{DBLP:conf/eccv/DaveKTSR20}
A.~Dave, T.~Khurana, P.~Tokmakov, C.~Schmid, and D.~Ramanan, ``{TAO:} {A}
  large-scale benchmark for tracking any object,'' in \emph{ECCV}, vol. 12350,
  2020, pp. 436--454.

\bibitem{DBLP:journals/ijcv/LiuOWFCLP20}
L.~Liu, W.~Ouyang, X.~Wang, P.~W. Fieguth, J.~Chen, X.~Liu, and
  M.~Pietik{\"{a}}inen, ``Deep learning for generic object detection: {A}
  survey,'' \emph{IJCV}, vol. 128, no.~2, pp. 261--318, 2020.

\bibitem{DBLP:journals/corr/abs-1912-00535}
S.~M. Marvasti{-}Zadeh, L.~Cheng, H.~Ghanei{-}Yakhdan, and S.~Kasaei, ``Deep
  learning for visual tracking: {A} comprehensive survey,'' \emph{IEEE
  Transactions on Intelligent Transportation Systems}, 2021.

\bibitem{DBLP:journals/ijon/CiaparroneSTTTH20}
G.~Ciaparrone, F.~L. S{\'{a}}nchez, S.~Tabik, L.~Troiano, R.~Tagliaferri, and
  F.~Herrera, ``Deep learning in video multi-object tracking: {A} survey,''
  \emph{Neurocomputing}, vol. 381, pp. 61--88, 2020.

\bibitem{DBLP:journals/pami/XuZYHRHS21}
X.~Xu, X.~Zhang, B.~Yu, X.~S. Hu, C.~Rowen, J.~Hu, and Y.~Shi, ``{DAC-SDC} low
  power object detection challenge for {UAV} applications,'' \emph{TPAMI},
  vol.~43, no.~2, pp. 392--403, 2021.

\bibitem{DBLP:journals/pami/DollarWSP12}
P.~Doll{\'{a}}r, C.~Wojek, B.~Schiele, and P.~Perona, ``Pedestrian detection:
  An evaluation of the state of the art,'' \emph{TPAMI}, vol.~34, no.~4, pp.
  743--761, 2012.

\bibitem{DBLP:journals/ijcv/EveringhamEGWWZ15}
M.~Everingham, S.~M.~A. Eslami, L.~J.~V. Gool, C.~K.~I. Williams, J.~M. Winn,
  and A.~Zisserman, ``The pascal visual object classes challenge: {A}
  retrospective,'' \emph{IJCV}, vol. 111, no.~1, pp. 98--136, 2015.

\bibitem{DBLP:conf/cvpr/PrestLCSF12}
A.~Prest, C.~Leistner, J.~Civera, C.~Schmid, and V.~Ferrari, ``Learning object
  class detectors from weakly annotated video,'' in \emph{CVPR}, 2012, pp.
  3282--3289.

\bibitem{DBLP:conf/avss/LyuCDWQLWKHCCAB17}
S.~Lyu, M.~Chang, D.~Du, L.~Wen, H.~Qi, Y.~Li, Y.~Wei, L.~Ke, T.~Hu, M.~D.
  Coco, P.~Carcagn{\`{\i}}, and \textit{et al.}, ``{UA-DETRAC} 2017: Report of
  {AVSS2017} {\&} {IWT4S} challenge on advanced traffic monitoring,'' in
  \emph{AVSS}, 2017, pp. 1--7.

\bibitem{DBLP:conf/avss/LyuCDLWCCSMDCBG18}
S.~Lyu, M.~Chang, D.~Du, W.~Li, Y.~Wei, M.~D. Coco, P.~Carcagn{\`{\i}}, and
  \textit{et al.}, ``{UA-DETRAC} 2018: Report of {AVSS2018} {\&} {IWT4S}
  challenge on advanced traffic monitoring,'' in \emph{AVSS}, 2018, pp. 1--6.

\bibitem{DBLP:journals/corr/abs-2003-09003}
P.~Dendorfer, H.~Rezatofighi, A.~Milan, J.~Shi, D.~Cremers, I.~D. Reid,
  S.~Roth, K.~Schindler, and L.~Leal{-}Taix{\'{e}}, ``{MOT20:} {A} benchmark
  for multi object tracking in crowded scenes,'' \emph{CoRR}, vol.
  abs/2003.09003, 2020.

\bibitem{DBLP:journals/pami/SmeuldersCCCDS14}
A.~W.~M. Smeulders, D.~M. Chu, R.~Cucchiara, S.~Calderara, A.~Dehghan, and
  M.~Shah, ``Visual tracking: An experimental survey,'' \emph{TPAMI}, vol.~36,
  no.~7, pp. 1442--1468, 2014.

\bibitem{DBLP:journals/tip/LiangBL15}
P.~Liang, E.~Blasch, and H.~Ling, ``Encoding color information for visual
  tracking: Algorithms and benchmark,'' \emph{TIP}, vol.~24, no.~12, pp.
  5630--5644, 2015.

\bibitem{DBLP:conf/iccv/GaloogahiFHRL17}
H.~K. Galoogahi, A.~Fagg, C.~Huang, D.~Ramanan, and S.~Lucey, ``Need for speed:
  {A} benchmark for higher frame rate object tracking,'' in \emph{ICCV}, 2017,
  pp. 1134--1143.

\bibitem{DBLP:conf/icra/LiangWLWLL18}
P.~Liang, Y.~Wu, H.~Lu, L.~Wang, C.~Liao, and H.~Ling, ``Planar object tracking
  in the wild: {A} benchmark,'' in \emph{ICRA}, 2018, pp. 651--658.

\bibitem{DBLP:conf/iccvw/KristanLMFPZVHL17}
M.~Kristan, A.~Leonardis, J.~Matas, M.~Felsberg, R.~P. Pflugfelder, L.~C. Zajc,
  T.~Vojir, G.~H{\"{a}}ger, A.~Lukezic, A.~Eldesokey, G.~Fern{\'{a}}ndez, and
  \textit{et al.}, ``The visual object tracking {VOT2017} challenge results,''
  in \emph{{ICCV}Workshops}, 2017, pp. 1949--1972.

\bibitem{DBLP:conf/eccv/KristanLMFPZVBL18}
M.~Kristan, A.~Leonardis, J.~Matas, M.~Felsberg, R.~P. Pflugfelder, L.~C. Zajc,
  T.~Voj{\'{\i}}r, G.~Bhat, A.~Lukezic, A.~Eldesokey, G.~Fern{\'{a}}ndez, and
  \textit{et al.}, ``The sixth visual object tracking {VOT2018} challenge
  results,'' in \emph{{ECCV}Workshops}, 2018, pp. 3--53.

\bibitem{DBLP:journals/tip/DuQLWHL16}
D.~Du, H.~Qi, W.~Li, L.~Wen, Q.~Huang, and S.~Lyu, ``Online deformable object
  tracking based on structure-aware hyper-graph,'' \emph{TIP}, vol.~25, no.~8,
  pp. 3572--3584, 2016.

\bibitem{DBLP:journals/ijcv/YuLZHDTS20}
H.~Yu, G.~Li, W.~Zhang, Q.~Huang, D.~Du, Q.~Tian, and N.~Sebe, ``The unmanned
  aerial vehicle benchmark: Object detection, tracking and baseline,''
  \emph{IJCV}, vol. 128, no.~5, pp. 1141--1159, 2020.

\bibitem{DBLP:journals/corr/abs-1810-10438}
Y.~Lyu, G.~Vosselman, G.~Xia, A.~Yilmaz, and M.~Y. Yang, ``The uavid dataset
  for video semantic segmentation,'' \emph{CoRR}, vol. abs/1810.10438, 2018.

\bibitem{DBLP:journals/pami/RenHG017}
S.~Ren, K.~He, R.~B. Girshick, and J.~Sun, ``Faster {R-CNN:} towards real-time
  object detection with region proposal networks,'' \emph{TPAMI}, vol.~39,
  no.~6, pp. 1137--1149, 2017.

\bibitem{DBLP:conf/iccv/HeGDG17}
K.~He, G.~Gkioxari, P.~Doll{\'{a}}r, and R.~B. Girshick, ``Mask {R-CNN},'' in
  \emph{ICCV}, 2017, pp. 2980--2988.

\bibitem{DBLP:conf/cvpr/CaiV18}
Z.~Cai and N.~Vasconcelos, ``Cascade {R-CNN:} delving into high quality object
  detection,'' in \emph{CVPR}, 2018, pp. 6154--6162.

\bibitem{DBLP:journals/corr/abs-1804-02767}
J.~Redmon and A.~Farhadi, ``Yolov3: An incremental improvement,'' \emph{CoRR},
  vol. abs/1804.02767, 2018.

\bibitem{DBLP:journals/corr/abs-1904-07850}
X.~Zhou, D.~Wang, and P.~Kr{\"{a}}henb{\"{u}}hl, ``Objects as points,''
  \emph{CoRR}, vol. abs/1904.07850, 2019.

\bibitem{DBLP:conf/cvpr/LinDGHHB17}
T.~Lin, P.~Doll{\'{a}}r, R.~B. Girshick, K.~He, B.~Hariharan, and S.~J.
  Belongie, ``Feature pyramid networks for object detection,'' in \emph{CVPR},
  2017, pp. 936--944.

\bibitem{DBLP:conf/eccv/LiuAESRFB16}
W.~Liu, D.~Anguelov, D.~Erhan, C.~Szegedy, S.~E. Reed, C.~Fu, and A.~C. Berg,
  ``{SSD:} single shot multibox detector,'' in \emph{ECCV}, 2016, pp. 21--37.

\bibitem{DBLP:conf/iccv/LinGGHD17}
T.~Lin, P.~Goyal, R.~B. Girshick, K.~He, and P.~Doll{\'{a}}r, ``Focal loss for
  dense object detection,'' in \emph{ICCV}, 2017, pp. 2999--3007.

\bibitem{DBLP:conf/cvpr/ZhangWBLL18}
S.~Zhang, L.~Wen, X.~Bian, Z.~Lei, and S.~Z. Li, ``Single-shot refinement
  neural network for object detection,'' in \emph{CVPR}, 2018, pp. 4203--4212.

\bibitem{DBLP:conf/iccv/Girshick15}
R.~B. Girshick, ``Fast {R-CNN},'' in \emph{ICCV}, 2015, pp. 1440--1448.

\bibitem{DBLP:journals/corr/abs-1711-07264}
Z.~Li, C.~Peng, G.~Yu, X.~Zhang, Y.~Deng, and J.~Sun, ``Light-head {R-CNN:} in
  defense of two-stage object detector,'' \emph{CoRR}, vol. abs/1711.07264,
  2017.

\bibitem{DBLP:conf/cvpr/RedmonF17}
J.~Redmon and A.~Farhadi, ``{YOLO9000:} better, faster, stronger,'' in
  \emph{CVPR}, 2017, pp. 6517--6525.

\bibitem{DBLP:journals/corr/abs-2004-10934}
A.~Bochkovskiy, C.~Wang, and H.~M. Liao, ``Yolov4: Optimal speed and accuracy
  of object detection,'' \emph{CoRR}, vol. abs/2004.10934, 2020.

\bibitem{DBLP:journals/corr/abs-2011-08036}
C.~Wang, A.~Bochkovskiy, and H.~M. Liao, ``Scaled-yolov4: Scaling cross stage
  partial network,'' in \emph{CVPR}, 2021, pp. 13\,029--13\,038.

\bibitem{DBLP:conf/iccv/DuanBXQH019}
K.~Duan, S.~Bai, L.~Xie, H.~Qi, Q.~Huang, and Q.~Tian, ``Centernet: Keypoint
  triplets for object detection,'' in \emph{ICCV}, 2019, pp. 6568--6577.

\bibitem{DBLP:journals/corr/abs-1903-00621}
C.~Zhu, Y.~He, and M.~Savvides, ``Feature selective anchor-free module for
  single-shot object detection,'' in \emph{CVPR}, 2019, pp. 840--849.

\bibitem{DBLP:journals/corr/abs-1904-01355}
Z.~Tian, C.~Shen, H.~Chen, and T.~He, ``{FCOS:} fully convolutional one-stage
  object detection,'' in \emph{ICCV}, 2019, pp. 9626--9635.

\bibitem{DBLP:conf/iccv/YangLHWL19}
Z.~Yang, S.~Liu, H.~Hu, L.~Wang, and S.~Lin, ``Reppoints: Point set
  representation for object detection,'' in \emph{ICCV}, 2019, pp. 9656--9665.

\bibitem{DBLP:conf/nips/Chen00WL020}
Y.~Chen, Z.~Zhang, Y.~Cao, L.~Wang, S.~Lin, and H.~Hu, ``Reppoints v2:
  Verification meets regression for object detection,'' in \emph{NeurIPS},
  2020.

\bibitem{DBLP:conf/aaai/LiuWWLZTL20}
Y.~Liu, Y.~Wang, S.~Wang, T.~Liang, Q.~Zhao, Z.~Tang, and H.~Ling, ``Cbnet: {A}
  novel composite backbone network architecture for object detection,'' in
  \emph{AAAI}, 2020, pp. 11\,653--11\,660.

\bibitem{DBLP:conf/cvpr/PangCSFOL19}
J.~Pang, K.~Chen, J.~Shi, H.~Feng, W.~Ouyang, and D.~Lin, ``Libra {R-CNN:}
  towards balanced learning for object detection,'' in \emph{CVPR}, 2019, pp.
  821--830.

\bibitem{DBLP:conf/cvpr/0009XLW19}
K.~Sun, B.~Xiao, D.~Liu, and J.~Wang, ``Deep high-resolution representation
  learning for human pose estimation,'' in \emph{CVPR}, 2019, pp. 5693--5703.

\bibitem{DBLP:conf/eccv/LiPYZDS18}
Z.~Li, C.~Peng, G.~Yu, X.~Zhang, Y.~Deng, and J.~Sun, ``Detnet: Design backbone
  for object detection,'' in \emph{ECCV}, 2018, pp. 339--354.

\bibitem{DBLP:journals/corr/abs-2006-02334}
S.~Qiao, L.~Chen, and A.~L. Yuille, ``Detectors: Detecting objects with
  recursive feature pyramid and switchable atrous convolution,'' in
  \emph{CVPR}, 2021, pp. 10\,213--10\,224.

\bibitem{DBLP:journals/corr/abs-1905-09646}
X.~Li, X.~Hu, and J.~Yang, ``Spatial group-wise enhance: Improving semantic
  feature learning in convolutional networks,'' \emph{CoRR}, vol.
  abs/1905.09646, 2019.

\bibitem{DBLP:journals/corr/abs-1904-11492}
Y.~Cao, J.~Xu, S.~Lin, F.~Wei, and H.~Hu, ``Gcnet: Non-local networks meet
  squeeze-excitation networks and beyond,'' in \emph{ICCV Workshops}.\hskip 1em
  plus 0.5em minus 0.4em\relax {IEEE}, 2019, pp. 1971--1980.

\bibitem{DBLP:conf/iccv/DaiQXLZHW17}
J.~Dai, H.~Qi, Y.~Xiong, Y.~Li, G.~Zhang, H.~Hu, and Y.~Wei, ``Deformable
  convolutional networks,'' in \emph{ICCV}, 2017, pp. 764--773.

\bibitem{DBLP:conf/cvpr/HuSS18}
J.~Hu, L.~Shen, and G.~Sun, ``Squeeze-and-excitation networks,'' in
  \emph{CVPR}, 2018.

\bibitem{DBLP:conf/iccv/WangCX0LL19}
J.~Wang, K.~Chen, R.~Xu, Z.~Liu, C.~C. Loy, and D.~Lin, ``{CARAFE:}
  content-aware reassembly of features,'' in \emph{ICCV}, 2019, pp. 3007--3016.

\bibitem{DBLP:journals/corr/abs-1901-03278}
J.~Wang, K.~Chen, S.~Yang, C.~C. Loy, and D.~Lin, ``Region proposal by guided
  anchoring,'' in \emph{CVPR}, 2019, pp. 2965--2974.

\bibitem{DBLP:conf/cvpr/ZhangCYLL20}
S.~Zhang, C.~Chi, Y.~Yao, Z.~Lei, and S.~Z. Li, ``Bridging the gap between
  anchor-based and anchor-free detection via adaptive training sample
  selection,'' in \emph{CVPR}, 2020, pp. 9756--9765.

\bibitem{DBLP:conf/eccv/KimL20}
K.~Kim and H.~S. Lee, ``Probabilistic anchor assignment with iou prediction for
  object detection,'' in \emph{ECCV}, vol. 12370, 2020, pp. 355--371.

\bibitem{DBLP:conf/nips/SinghND18}
B.~Singh, M.~Najibi, and L.~S. Davis, ``{SNIPER:} efficient multi-scale
  training,'' in \emph{NeurIPS}, 2018, pp. 9333--9343.

\bibitem{DBLP:conf/cvpr/YuWPGYS18}
C.~Yu, J.~Wang, C.~Peng, C.~Gao, G.~Yu, and N.~Sang, ``Learning a
  discriminative feature network for semantic segmentation,'' in \emph{CVPR},
  2018, pp. 1857--1866.

\bibitem{DBLP:conf/cvpr/ZhangQX0WY18}
Z.~Zhang, S.~Qiao, C.~Xie, W.~Shen, B.~Wang, and A.~L. Yuille, ``Single-shot
  object detection with enriched semantics,'' in \emph{CVPR}, 2018, pp.
  5813--5821.

\bibitem{DBLP:conf/iccv/YangFCBL19}
F.~Yang, H.~Fan, P.~Chu, E.~Blasch, and H.~Ling, ``Clustered object detection
  in aerial images,'' in \emph{ICCV}, 2019, pp. 8310--8319.

\bibitem{DBLP:conf/iccv/NajibiS019}
M.~Najibi, B.~Singh, and L.~Davis, ``Autofocus: Efficient multi-scale
  inference,'' in \emph{ICCV}, 2019, pp. 9744--9754.

\bibitem{DBLP:conf/eccv/LawD18}
H.~Law and J.~Deng, ``Cornernet: Detecting objects as paired keypoints,'' in
  \emph{ECCV}, 2018, pp. 765--781.

\bibitem{DBLP:conf/cvpr/WangJQYLZWT17}
F.~Wang, M.~Jiang, C.~Qian, S.~Yang, C.~Li, H.~Zhang, X.~Wang, and X.~Tang,
  ``Residual attention network for image classification,'' in \emph{CVPR},
  2017, pp. 6450--6458.

\bibitem{DBLP:conf/cvpr/HarisSU18}
M.~Haris, G.~Shakhnarovich, and N.~Ukita, ``Deep back-projection networks for
  super-resolution,'' in \emph{CVPR}, 2018, pp. 1664--1673.

\bibitem{DBLP:conf/nips/TarvainenV17}
A.~Tarvainen and H.~Valpola, ``Mean teachers are better role models:
  Weight-averaged consistency targets improve semi-supervised deep learning
  results,'' in \emph{NeurIPS}, 2017, pp. 1195--1204.

\bibitem{DBLP:conf/aaai/CaiWZDW21}
Y.~Cai, L.~Wen, L.~Zhang, D.~Du, and W.~Wang, ``Rethinking object detection in
  retail stores,'' in \emph{AAAI}, 2021, pp. 947--954.

\bibitem{DBLP:conf/cvpr/DanelljanBKF17}
M.~Danelljan, G.~Bhat, F.~S. Khan, and M.~Felsberg, ``{ECO:} efficient
  convolution operators for tracking,'' in \emph{CVPR}, 2017, pp. 6931--6939.

\bibitem{DBLP:journals/corr/abs-1812-11703}
B.~Li, W.~Wu, Q.~Wang, F.~Zhang, J.~Xing, and J.~Yan, ``Siamrpn++: Evolution of
  siamese visual tracking with very deep networks,'' in \emph{CVPR}, 2018.

\bibitem{DBLP:journals/corr/KangLYZYXZWWWO16}
K.~Kang, H.~Li, J.~Yan, X.~Zeng, B.~Yang, T.~Xiao, C.~Zhang, Z.~Wang, R.~Wang,
  X.~Wang, and W.~Ouyang, ``{T-CNN:} tubelets with convolutional neural
  networks for object detection from videos,'' \emph{TCSVT}, vol.~28, no.~10,
  pp. 2896--2907, 2018.

\bibitem{DBLP:journals/corr/HanKPRBSLYH16}
W.~Han, P.~Khorrami, T.~L. Paine, P.~Ramachandran, M.~Babaeizadeh, H.~Shi,
  J.~Li, S.~Yan, and T.~S. Huang, ``Seq-nms for video object detection,''
  \emph{CoRR}, vol. abs/1602.08465, 2016.

\bibitem{DBLP:conf/iccv/FeichtenhoferPZ17}
C.~Feichtenhofer, A.~Pinz, and A.~Zisserman, ``Detect to track and track to
  detect,'' in \emph{ICCV}, 2017, pp. 3057--3065.

\bibitem{DBLP:conf/iccv/ZhuWDYW17}
X.~Zhu, Y.~Wang, J.~Dai, L.~Yuan, and Y.~Wei, ``Flow-guided feature aggregation
  for video object detection,'' in \emph{ICCV}, 2017, pp. 408--417.

\bibitem{DBLP:conf/cvpr/ZhuXDYW17}
X.~Zhu, Y.~Xiong, J.~Dai, L.~Yuan, and Y.~Wei, ``Deep feature flow for video
  recognition,'' in \emph{CVPR}, 2017, pp. 4141--4150.

\bibitem{DBLP:conf/eccv/WangZYD18}
S.~Wang, Y.~Zhou, J.~Yan, and Z.~Deng, ``Fully motion-aware network for video
  object detection,'' in \emph{ECCV}, vol. 11217, 2018, pp. 557--573.

\bibitem{DBLP:conf/cvpr/Chen0HW20}
Y.~Chen, Y.~Cao, H.~Hu, and L.~Wang, ``Memory enhanced global-local aggregation
  for video object detection,'' in \emph{CVPR}, 2020, pp. 10\,334--10\,343.

\bibitem{DBLP:conf/iccv/LuLT17}
Y.~Lu, C.~Lu, and C.~Tang, ``Online video object detection using association
  {LSTM},'' in \emph{ICCV}, 2017, pp. 2363--2371.

\bibitem{DBLP:conf/eccv/XiaoL18}
F.~Xiao and Y.~J. Lee, ``Video object detection with an aligned
  spatial-temporal memory,'' in \emph{ECCV}, vol. 11212, 2018, pp. 494--510.

\bibitem{DBLP:conf/iccv/DengHSZXMRG19}
H.~Deng, Y.~Hua, T.~Song, Z.~Zhang, Z.~Xue, R.~Ma, N.~M. Robertson, and
  H.~Guan, ``Object guided external memory network for video object
  detection,'' in \emph{ICCV}, 2019, pp. 6677--6686.

\bibitem{DBLP:conf/cvpr/BertinettoVGMT16}
L.~Bertinetto, J.~Valmadre, S.~Golodetz, O.~Miksik, and P.~H.~S. Torr,
  ``Staple: Complementary learners for real-time tracking,'' in \emph{CVPR},
  2016, pp. 1401--1409.

\bibitem{DBLP:conf/cvpr/MuellerSG17}
M.~Mueller, N.~Smith, and B.~Ghanem, ``Context-aware correlation filter
  tracking,'' in \emph{CVPR}, 2017, pp. 1387--1395.

\bibitem{DBLP:conf/eccv/DanelljanRKF16}
M.~Danelljan, A.~Robinson, F.~S. Khan, and M.~Felsberg, ``Beyond correlation
  filters: Learning continuous convolution operators for visual tracking,'' in
  \emph{ECCV}, vol. 9909, 2016, pp. 472--488.

\bibitem{DBLP:conf/iccvw/HeFZDB17}
Z.~He, Y.~Fan, J.~Zhuang, Y.~Dong, and H.~Bai, ``Correlation filters with
  weighted convolution responses,'' in \emph{ICCVWorkshops}, 2017, pp.
  1992--2000.

\bibitem{DBLP:conf/iccv/GaloogahiFL17}
H.~K. Galoogahi, A.~Fagg, and S.~Lucey, ``Learning background-aware correlation
  filters for visual tracking,'' in \emph{ICCV}, 2017, pp. 1144--1152.

\bibitem{DBLP:conf/eccv/BertinettoVHVT16}
L.~Bertinetto, J.~Valmadre, J.~F. Henriques, A.~Vedaldi, and P.~H.~S. Torr,
  ``Fully-convolutional siamese networks for object tracking,'' in \emph{ECCV},
  2016, pp. 850--865.

\bibitem{DBLP:conf/iccv/Guo0ZHWW17}
Q.~Guo, W.~Feng, C.~Zhou, R.~Huang, L.~Wan, and S.~Wang, ``Learning dynamic
  siamese network for visual object tracking,'' in \emph{ICCV}, 2017, pp.
  1781--1789.

\bibitem{DBLP:journals/corr/abs-1812-05050}
Q.~Wang, L.~Zhang, L.~Bertinetto, W.~Hu, and P.~H.~S. Torr, ``Fast online
  object tracking and segmentation: {A} unifying approach,'' in \emph{CVPR},
  2019, pp. 1328--1338.

\bibitem{DBLP:conf/cvpr/VoigtlaenderLTL20}
P.~Voigtlaender, J.~Luiten, P.~H.~S. Torr, and B.~Leibe, ``Siam {R-CNN:} visual
  tracking by re-detection,'' in \emph{CVPR}, 2020, pp. 6577--6587.

\bibitem{DBLP:conf/cvpr/NamH16}
H.~Nam and B.~Han, ``Learning multi-domain convolutional neural networks for
  visual tracking,'' in \emph{CVPR}, 2016, pp. 4293--4302.

\bibitem{DBLP:journals/corr/abs-1804-04273}
Y.~Song, C.~Ma, X.~Wu, L.~Gong, L.~Bao, W.~Zuo, C.~Shen, R.~W.~H. Lau, and
  M.~Yang, ``{VITAL:} visual tracking via adversarial learning,'' in
  \emph{CVPR}, 2018.

\bibitem{DBLP:conf/cvpr/ValmadreBHVT17}
J.~Valmadre, L.~Bertinetto, J.~F. Henriques, A.~Vedaldi, and P.~H.~S. Torr,
  ``End-to-end representation learning for correlation filter based tracking,''
  in \emph{CVPR}, 2017, pp. 5000--5008.

\bibitem{DBLP:conf/cvpr/ChoiC0YLJD018}
J.~Choi, H.~J. Chang, T.~Fischer, S.~Yun, K.~Lee, J.~Jeong, Y.~Demiris, and
  J.~Y. Choi, ``Context-aware deep feature compression for high-speed visual
  tracking,'' in \emph{CVPR}, 2018, pp. 479--488.

\bibitem{DBLP:journals/corr/abs-1811-07628}
M.~Danelljan, G.~Bhat, F.~S. Khan, and M.~Felsberg, ``{ATOM:} accurate tracking
  by overlap maximization,'' in \emph{CVPR}, 2019, pp. 4660--4669.

\bibitem{DBLP:conf/iccv/BhatDGT19}
G.~Bhat, M.~Danelljan, L.~V. Gool, and R.~Timofte, ``Learning discriminative
  model prediction for tracking,'' in \emph{ICCV}, 2019, pp. 6181--6190.

\bibitem{DBLP:conf/cvpr/RealSMPV17}
E.~Real, J.~Shlens, S.~Mazzocchi, X.~Pan, and V.~Vanhoucke,
  ``Youtube-boundingboxes: {A} large high-precision human-annotated data set
  for object detection in video,'' in \emph{CVPR}, 2017, pp. 7464--7473.

\bibitem{DBLP:conf/iccvw/YingLRWW17}
Z.~Ying, G.~Li, Y.~Ren, R.~Wang, and W.~Wang, ``A new low-light image
  enhancement algorithm using camera response model,'' in \emph{{ICCV}
  Workshops}, 2017, pp. 3015--3022.

\bibitem{DBLP:conf/eccv/BhatJDKF18}
G.~Bhat, J.~Johnander, M.~Danelljan, F.~S. Khan, and M.~Felsberg, ``Unveiling
  the power of deep tracking,'' in \emph{ECCV}, 2018, pp. 493--509.

\bibitem{DBLP:conf/eccv/ZhuWLWYH18}
Z.~Zhu, Q.~Wang, B.~Li, W.~Wu, J.~Yan, and W.~Hu, ``Distractor-aware siamese
  networks for visual object tracking,'' in \emph{ECCV}, 2018, pp. 103--119.

\bibitem{DBLP:conf/icip/WojkeBP17}
N.~Wojke, A.~Bewley, and D.~Paulus, ``Simple online and realtime tracking with
  a deep association metric,'' in \emph{ICIP}, 2017, pp. 3645--3649.

\bibitem{DBLP:journals/pami/KalalMM12}
Z.~Kalal, K.~Mikolajczyk, and J.~Matas, ``Tracking-learning-detection,''
  \emph{TPAMI}, vol.~34, no.~7, pp. 1409--1422, 2012.

\bibitem{isvrc-2017}
E.~Park, W.~Liu, O.~Russakovsky, J.~Deng, F.-F. Li, and A.~Berg, ``{L}arge
  {S}cale {V}isual {R}ecognition {C}hallenge 2017,''
  http://image-net.org/challenges/LSVRC/2017, 2017.

\bibitem{DBLP:journals/pami/HenriquesC0B15}
J.~F. Henriques, R.~Caseiro, P.~Martins, and J.~Batista, ``High-speed tracking
  with kernelized correlation filters,'' \emph{TPAMI}, vol.~37, no.~3, pp.
  583--596, 2015.

\bibitem{DBLP:conf/icip/BewleyGORU16}
A.~Bewley, Z.~Ge, L.~Ott, F.~T. Ramos, and B.~Upcroft, ``Simple online and
  realtime tracking,'' in \emph{ICIP}, 2016, pp. 3464--3468.

\bibitem{DBLP:conf/cvpr/PirsiavashRF11}
H.~Pirsiavash, D.~Ramanan, and C.~C. Fowlkes, ``Globally-optimal greedy
  algorithms for tracking a variable number of objects,'' in \emph{CVPR}, 2011,
  pp. 1201--1208.

\bibitem{DBLP:conf/avss/BochinskiES17}
E.~Bochinski, V.~Eiselein, and T.~Sikora, ``High-speed tracking-by-detection
  without using image information,'' in \emph{AVSS}, 2017, pp. 1--6.

\bibitem{DBLP:conf/icmcs/ChenAZS18}
L.~Chen, H.~Ai, Z.~Zhuang, and C.~Shang, ``Real-time multiple people tracking
  with deeply learned candidate selection and person re-identification,'' in
  \emph{ICME}, 2018, pp. 1--6.

\bibitem{DBLP:journals/corr/abs-1811-07258}
G.~Wang, Y.~Wang, H.~Zhang, R.~Gu, and J.~Hwang, ``Exploit the connectivity:
  Multi-object tracking with trackletnet,'' in \emph{ACM MM}, 2019.

\bibitem{DBLP:conf/iccv/BergmannML19}
P.~Bergmann, T.~Meinhardt, and L.~Leal{-}Taix{\'{e}}, ``Tracking without bells
  and whistles,'' in \emph{ICCV}, 2019, pp. 941--951.

\bibitem{DBLP:journals/corr/abs-2004-01888}
Y.~Zhang, C.~Wang, X.~Wang, W.~Zeng, and W.~Liu, ``Fairmot: On the fairness of
  detection and re-identification in multiple object tracking,'' \emph{IJCV},
  2021.

\bibitem{DBLP:conf/eccv/ZhouKK20}
X.~Zhou, V.~Koltun, and P.~Kr{\"{a}}henb{\"{u}}hl, ``Tracking objects as
  points,'' in \emph{ECCV}, vol. 12349, 2020, pp. 474--490.

\bibitem{DBLP:journals/corr/abs-1810-11780}
S.~Sun, N.~Akhtar, H.~Song, A.~Mian, and M.~Shah, ``Deep affinity network for
  multiple object tracking,'' \emph{TPAMI}, 2019.

\bibitem{DBLP:conf/iccv/ZhouYCX19}
K.~Zhou, Y.~Yang, A.~Cavallaro, and T.~Xiang, ``Omni-scale feature learning for
  person re-identification,'' in \emph{ICCV}, 2019, pp. 3701--3711.

\bibitem{DBLP:conf/cvpr/0004GLL019}
H.~Luo, Y.~Gu, X.~Liao, S.~Lai, and W.~Jiang, ``Bag of tricks and a strong
  baseline for deep person re-identification,'' in \emph{CVPRW}, 2019, pp.
  1487--1495.

\bibitem{DBLP:conf/mm/WangYCLZ18}
G.~Wang, Y.~Yuan, X.~Chen, J.~Li, and X.~Zhou, ``Learning discriminative
  features with multiple granularities for person re-identification,'' in
  \emph{ACM MM}, 2018, pp. 274--282.

\bibitem{DBLP:conf/icpr/KalalMM10}
Z.~Kalal, K.~Mikolajczyk, and J.~Matas, ``Forward-backward error: Automatic
  detection of tracking failures,'' in \emph{ICPR}, 2010, pp. 2756--2759.

\bibitem{DBLP:conf/cvpr/IlgMSKDB17}
E.~Ilg, N.~Mayer, T.~Saikia, M.~Keuper, A.~Dosovitskiy, and T.~Brox, ``Flownet
  2.0: Evolution of optical flow estimation with deep networks,'' in
  \emph{CVPR}, 2017, pp. 1647--1655.

\bibitem{DBLP:conf/cvpr/SunY0K18}
D.~Sun, X.~Yang, M.~Liu, and J.~Kautz, ``Pwc-net: Cnns for optical flow using
  pyramid, warping, and cost volume,'' in \emph{CVPR}, 2018, pp. 8934--8943.

\bibitem{DBLP:conf/aaai/MilanRD0S17}
A.~Milan, S.~H. Rezatofighi, A.~R. Dick, I.~D. Reid, and K.~Schindler, ``Online
  multi-target tracking using recurrent neural networks,'' in \emph{AAAI},
  2017, pp. 4225--4232.

\bibitem{DBLP:conf/iccv/SadeghianAS17}
A.~Sadeghian, A.~Alahi, and S.~Savarese, ``Tracking the untrackable: Learning
  to track multiple cues with long-term dependencies,'' in \emph{ICCV}, 2017,
  pp. 300--311.

\bibitem{DBLP:conf/cvpr/ViolaJ01}
P.~A. Viola and M.~J. Jones, ``Rapid object detection using a boosted cascade
  of simple features,'' in \emph{CVPR}, 2001, pp. 511--518.

\bibitem{DBLP:conf/cvpr/FelzenszwalbMR08}
P.~F. Felzenszwalb, D.~A. McAllester, and D.~Ramanan, ``A discriminatively
  trained, multiscale, deformable part model,'' in \emph{CVPR}, 2008.

\bibitem{DBLP:journals/pami/DollarABP14}
P.~Doll{\'{a}}r, R.~Appel, S.~J. Belongie, and P.~Perona, ``Fast feature
  pyramids for object detection,'' \emph{TPAMI}, vol.~36, no.~8, pp.
  1532--1545, 2014.

\bibitem{DBLP:conf/eccv/HuangWN08}
C.~Huang, B.~Wu, and R.~Nevatia, ``Robust object tracking by hierarchical
  association of detection responses,'' in \emph{ECCV}, vol. 5303, 2008, pp.
  788--801.

\bibitem{DBLP:conf/bmvc/WangF15}
S.~Wang and C.~C. Fowlkes, ``Learning optimal parameters for multi-target
  tracking,'' in \emph{BMVC}, 2015, pp. 4.1--4.13.

\bibitem{DBLP:conf/aaai/WenDLBL19}
L.~Wen, D.~Du, S.~Li, X.~Bian, and S.~Lyu, ``Learning non-uniform hypergraph
  for multi-object tracking,'' in \emph{AAAI}, 2019, pp. 8981--8988.

\bibitem{DBLP:conf/cvpr/VoigtlaenderKOL19}
P.~Voigtlaender, M.~Krause, A.~Osep, J.~Luiten, B.~B.~G. Sekar, A.~Geiger, and
  B.~Leibe, ``{MOTS:} multi-object tracking and segmentation,'' in \emph{CVPR},
  2019, pp. 7942--7951.

\bibitem{DBLP:conf/cvpr/LuRVH20}
Z.~Lu, V.~Rathod, R.~Votel, and J.~Huang, ``Retinatrack: Online single stage
  joint detection and tracking,'' in \emph{CVPR}, 2020, pp. 14\,656--14\,666.

\bibitem{DBLP:conf/cvpr/PangLZLL20}
B.~Pang, Y.~Li, Y.~Zhang, M.~Li, and C.~Lu, ``Tubetk: Adopting tubes to track
  multi-object in a one-step training model,'' in \emph{CVPR}, 2020, pp.
  6307--6317.

\bibitem{DBLP:journals/corr/abs-1911-13053}
C.~Li, J.~Peng, L.~Yuan, G.~Wang, X.~Liang, L.~Lin, and X.~Chang, ``Blockwisely
  supervised neural architecture search with knowledge distillation,''
  \emph{CoRR}, vol. abs/1911.13053, 2019.

\bibitem{DBLP:conf/cvpr/YanDZTWF21}
Z.~Yan, X.~Dai, P.~Zhang, Y.~Tian, B.~Wu, and M.~Feiszli, ``{FP-NAS:} fast
  probabilistic neural architecture search,'' in \emph{CVPR}, 2021, pp.
  15\,139--15\,148.

\bibitem{DBLP:conf/cvpr/Collins12}
R.~T. Collins, ``Multitarget data association with higher-order motion
  models,'' in \emph{CVPR}, 2012, pp. 1744--1751.

\bibitem{DBLP:conf/cvpr/ShiLHYX14}
X.~Shi, H.~Ling, W.~Hu, C.~Yuan, and J.~Xing, ``Multi-target tracking with
  motion context in tensor power iteration,'' in \emph{CVPR}, 2014, pp.
  3518--3525.

\end{thebibliography}
}


\end{document}